\documentclass{article} 
\usepackage{arxiv_style,times}

\usepackage{amsmath,amsfonts,bm}

\def\eqref#1{equation~\ref{#1}}

\def\1{\bm{1}}

\DeclareMathAlphabet{\mathsfit}{\encodingdefault}{\sfdefault}{m}{sl}
\SetMathAlphabet{\mathsfit}{bold}{\encodingdefault}{\sfdefault}{bx}{n}

\usepackage[table]{xcolor}
\usepackage{hyperref}
\usepackage{url}
\usepackage{hyperref}
\usepackage{amsmath}
\usepackage{amssymb}
\usepackage{booktabs}
\usepackage{graphicx}
\usepackage{xcolor}
\usepackage{multirow}
\usepackage{microtype}
\usepackage{float}
\usepackage{rotating}
\usepackage[percent]{overpic}
\usepackage{makecell}
\usepackage{tcolorbox}
\usepackage{enumitem}
\usepackage{tikz}
\usepackage{arydshln}
\usepackage{wrapfig,needspace}
\usepackage{titlesec}
\setlist[itemize]{
    label={},
    leftmargin=0pt,
    itemsep=2pt,
    topsep=2pt,
    parsep=0pt,
    partopsep=0pt
}
\usepackage{multirow}
\usepackage[table]{xcolor}
\usepackage{makecell}
\newcommand{\Hrgb}{H_{\mathrm{rgb}}}
\newcommand{\Htac}{H_{\mathrm{tac}}}
\newcommand{\woevent}{{w/o Events}}
\newcommand{\film}{{FiLM}}

\newcommand{\rewind}{{ReWiND}}
\newcommand{\robometer}{{Robometer}}
\newcommand{\kone}{\AlgName{} \small{(\emph{K=1})}}
\newcommand{\kthree}{\AlgName{} \small{(\emph{K=3})}}
\newcommand{\htwo}{\AlgName{} \small{({h=2})}}
\newcommand{\inspectone}{\emph{Inspect Box 1}}
\newcommand{\inspecttwo}{"\emph{Inspect Box 2}"}

\newcommand{\contprog}{%
\raisebox{-0.25ex}{\resizebox{0.55cm}{!}{%
\begin{tikzpicture}
  \useasboundingbox (0,-0.55) rectangle (3.2,1.1);
  \draw[gray!35,->] (0,0) -- (3.2,0);
  \draw[gray!35,->] (0,0) -- (0,1.1);
  \draw[very thick] (0,0) -- (2.8,0.9);
\end{tikzpicture}}}
}

\newcommand{\discprog}{%
\raisebox{-0.25ex}{\resizebox{0.55cm}{!}{%
\begin{tikzpicture}
  \useasboundingbox (0,-0.55) rectangle (3.2,1.1);
  \draw[gray!35,->] (0,0) -- (3.2,0);
  \draw[gray!35,->] (0,0) -- (0,1.1);
  \draw[very thick] (0,0) -- (2.0,0) -- (2.0,0.9) -- (2.8,0.9);
\end{tikzpicture}}}
}
\newcommand{\condprog}{%
\raisebox{-0.25ex}{\resizebox{0.55cm}{!}{%
\begin{tikzpicture}
  \useasboundingbox (0,-1.1) rectangle (3.2,1.1);
  \draw[gray!35,->] (0,0) -- (3.2,0);
  \draw[gray!35,->] (0,-1.0) -- (0,1.1);
  \draw[very thick] (0,0) -- (2.0,0);
  \draw[very thick] (2.0,0) -- (2.0,0.9) -- (2.8,0.9);
  \draw[very thick,red] (2.0,0) -- (2.0,-0.9) -- (2.8,-0.9);
\end{tikzpicture}}}
}

\title{What to Attend, What to Keep: \\Skill-Conditioned Visuotactile Representation with Progress-Guided Event Memory}

\author{\begin{tabular}{l}
\\
\\
Amir-Hossein Shahidzadeh$^{1}$, Seungjae Lee$^{1}$, Eadom Dessalene$^{1}$, Shanthosh Mageswari$^{1}$,\\
Soroush Etemad$^{1}$, Furong Huang$^{1}$, Cornelia Fermüller$^{1}$, Yiannis Aloimonos$^{1}$\\[1ex]
{\normalfont\mdseries $^{1}$ Computer Science Department, University of Maryland, College Park, MD, USA}
\end{tabular}}

\newcommand{\AlgName}{\textsc{\textcolor[rgb]{0.47,0.134,0.188}{\textbf{S}kill\textbf{F}ormer}}}
\usepackage{xcolor}

\makeatletter
\definecolor{royalred}{rgb}{0.47,0.134,0.188}
\titleformat{\paragraph}[runin]
  {\normalfont\normalsize\bfseries\color[rgb]{0.47,0.134,0.188}}
  {}
  {0pt}
  {}

\titlespacing*{\paragraph}
  {0pt}  
  {0pt}  
  {5pt}  
\begin{document}

\maketitle
\begin{abstract}
Robotic manipulation integrates vision, touch, and language, whose importance
shifts across stages: vision guides reaching, while touch, through its
evolution over time, decides grasping, alignment, and contact. Yet existing multi-modal manipulation policies typically use fixed temporal
contexts and fusion strategies, despite shifts in what each
modality contributes across different skills. We
study how vision and touch should be combined at the level of primitive
skills, asking what each skill needs from each sensor, and propose a
skill-conditioned representation in which the queried skill conditions fusion over modality-specific short-term observation tokens while attending to a sparse event memory that retains terminal observations from the last $K$ executed skills. Evaluated by
skill progress estimation on three contact-rich tasks, it reduces
slip-detection delay by 87\% against fine-tuned SOTA progress models,
twist-completion delay by 67.5\% against a vision-only ablation, and progress error on a
blind search task by 92\% through sparse event memory. Gains concentrate
exactly where completion is defined by contact or task history. More broadly,
our results suggest that observation formation—not only policy architecture—is
a central challenge in multi-modal representation. \href{http://what-to-attend-what-to-keep.github.io/}{Project Website: http://what-to-attend-what-to-keep.github.io/}
\end{abstract}

\begin{figure}[h!]
  \centering
  \vspace{-0.3cm}
  \includegraphics[width=0.9\linewidth]{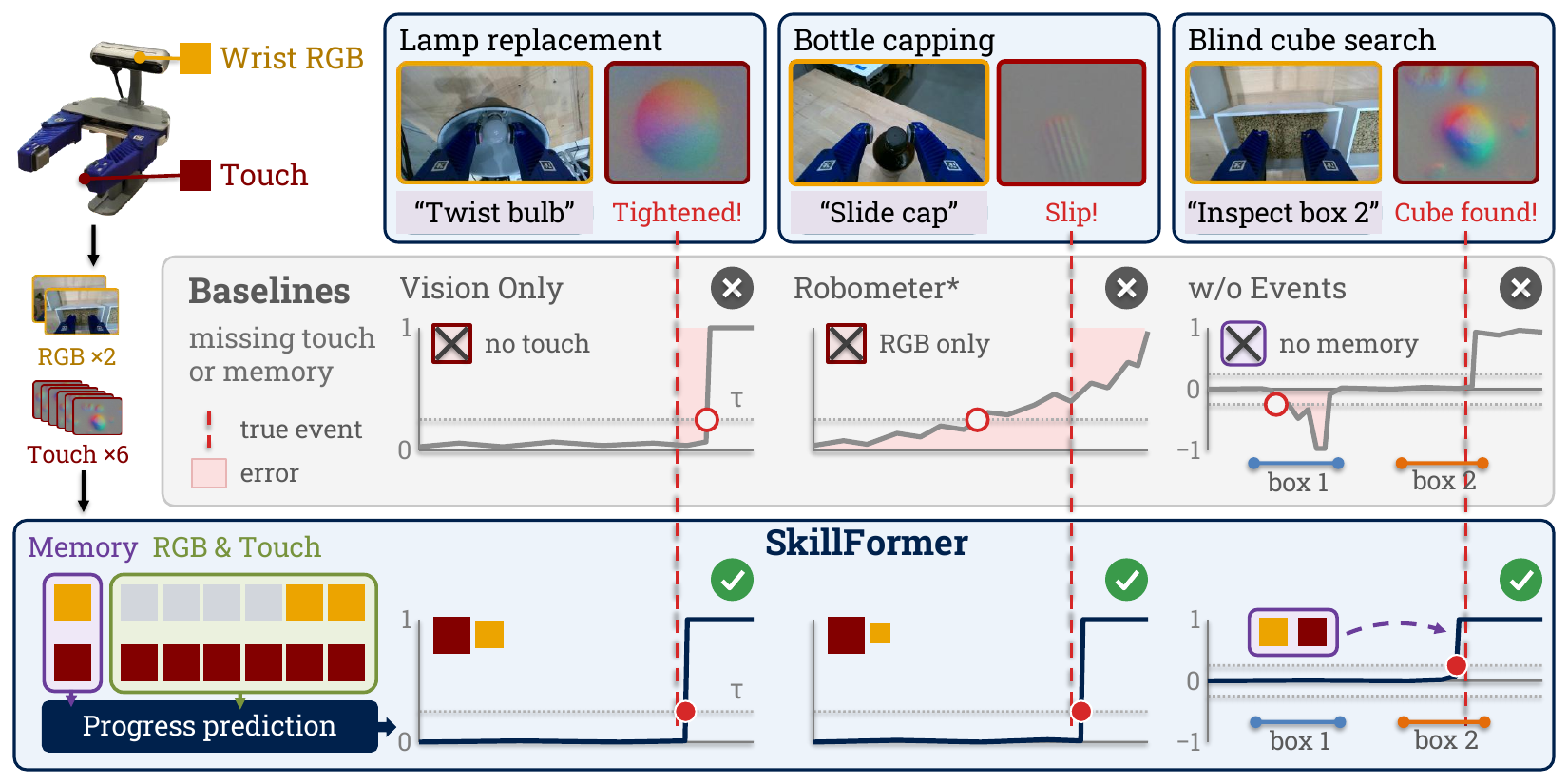
  }
  \vspace{-0.3cm}
  \caption{\textbf{\AlgName{} learns what to attend to and what to remember.} 
  Skill-conditioned fusion combines visual and tactile observation windows 
  with progress-guided event memory. Tactile observations reveal tightening 
  and slip events that are difficult to distinguish visually, while memory 
  helps distinguish repeated inspections across boxes. By connecting event 
  detection with retention, \AlgName{} turns brief sensory observations into 
  lasting context for interpreting subsequent interactions.}

  \label{fig:banner}
\end{figure} 
\newpage
\section{Introduction}
Robotic manipulation must prioritize different modalities at different stages of a
task \cite{lee2019making,chen2022visuo}. Vision provides spatial context for locating objects and guiding a reach,
whereas tactile feedback helps distinguish contact states during grasping,
alignment, and insertion \cite{huang20243d}. The relevant observations also span different time
scales: assessing bulb twisting requires interpreting how tactile readings
evolve, while detecting bottle-cap slip hinges on recognizing a brief event.
Similar skills can also produce distinct tactile patterns: aligning a bulb
with socket threads and aligning a cap with bottle threads need not share the
same contact signature. A useful representation must therefore learn which
sensory patterns and temporal changes matter for each skill.

Despite these changing requirements, common observation formulations use fixed,
short histories. Diffusion Policy uses short observation
windows~\citep{chi2023diffusion}, while the VLA models condition on a
single image and a language instruction~\citep{yu2026dm0embodiednativevisionlanguageactionmodel}. Applying the
same multimodal windows and fusion structure across skills leaves their
different sensory needs implicit. Short windows can also omit earlier events:
when searching successive containers, the current view may not reveal which
ones have already been inspected. Simply extending history is insufficient.
Dense visual patch tokens increase attention cost quadratically \cite{vaswani2017attention} with sequence
length under standard self-attention, while consecutive frames often repeat
similar information. This abundance of visual tokens risks obscuring the
sparse tactile changes that distinguish contact states.

The challenge is therefore to organize observations, not merely accumulate
them. Language-based interaction offers a useful analogy: dialogue turns and
tool calls provide explicit events around which memory can be organized.
Manipulation instead produces continuous sensory streams, whose meaningful
physical transitions must be inferred over time. Language-described primitive
skills identify the interaction being queried, while progress supervision
connects these descriptions to sensory evidence of completion. 



We introduce \AlgName{}, a language-conditioned visuotactile representation
combining modality-specialized encoders, asymmetric short-term horizons, and
sparse event-triggered memory. Fixed modality-specific windows capture recent visual and contact dynamics; 
\begin{wrapfigure}{r}{0.38\textwidth}
\vspace{-8pt}
\centering
\includegraphics[width=\linewidth]{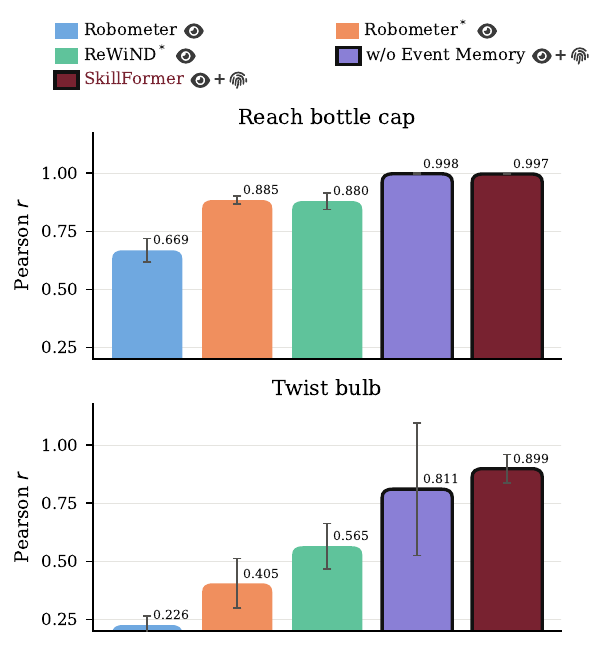}
\vspace{-0.8 cm}
\caption{\footnotesize\raggedright
 Pearson $r$ measures correlation between predicted and ground-truth progress (higher is better). Greater variability on twisting \emph{w/o event memory} suggests less consistent performance that does not establish online generalization. \small{$^*$: fine-tuned; eye/fingerprint: vision/touch.}}
\label{fig:intro_progress}
\vspace{-8pt}
\end{wrapfigure}
event memory retains sensory windows at predicted
skill completion for use beyond those horizons. A language query conditions
joint attention over both sources. At skill transitions, the query
and progress objective change together, encouraging phase-specific
representations within a shared network. This formulation makes
\emph{observation formation}---what to attend to, and what to
keep---a central design question alongside policy architecture.

\paragraph{Progress estimation as the lens.}
Skill progress provides structured, interpretable supervision for representing
geometric state, contact dynamics, and interaction history without jointly
optimizing an action policy.
Robometer and ReWiND demonstrate the value of learned progress and reward
signals for robot learning~\citep{robometer2025,rewind2025}. Motivated by these
works, we use this objective to probe task-relevant information in latent
representations. Queries for
active and inactive skills test whether the representation distinguishes
completed, ongoing, and not-yet-started skills. Across lamp replacement, bottle capping, and blind object search, we measure how skill-progress prediction changes when tactile input or event memory is removed, observation histories are shortened, or the fusion mechanism is changed. Section~\ref{sec:memory} specifies the
event-construction protocol and its annotation access.

Our primary contributions are as follows:
\begin{itemize}[leftmargin=1.2em]
    \item \textbf{A framework for visuotactile representation learning:}
    We introduce \AlgName{}, which uses language-described skills to
    organize perception and memory for contact-rich manipulation.

    \item \textbf{Skill-conditioned fusion of vision and touch:}
    We introduce a skill-conditioned attention mechanism that integrates
    visual and tactile observations. Our model reduces overall
    skill-progress MSE by 37--80\% compared with vision alone across
    the three manipulation tasks.

    \item \textbf{Sparse memory of past interactions:}
    We use predicted skill completions to form a sparse memory of earlier
    interactions. Event memory reduces overall skill-progress MSE by
    59--92\% across all three tasks compared with using recent observations alone.

    \item \textbf{Three contact-rich manipulation datasets:}
    We collect 429 demonstrations of lamp replacement, bottle capping,
    and blind cube search, pairing synchronized video and tactile
    recordings with primitive skill annotations.
\end{itemize}
\section{Background}
\paragraph{Multimodal representations for manipulation.}
Visuotactile learning combines scene-level visual information with local contact
observations~\citep{lee2019making,li2022see}. The usefulness of each modality
depends on the interaction, motivating a study of its feature space,
temporal context, and role in fusion \cite{feng2025anytouch,higuera2024sparsh}. We build on CLIP visual and language
features~\citep{radford2021learning} and the tactile-specialized FeelAnyForce
encoder~\citep{Shahidzadeh2024FeelAnyForceEC}. A generic DINOv2 encoder
\citep{oquab2023dinov2} provides a control for tactile specialization.
The contribution studied here is how these features and their histories are
organized around skills, beyond the selection of a pretrained backbone.

\paragraph{Skill conditioning and temporal representations.}
Language-described skills provide an interface for structured manipulation
\citep{kroemer2021review,ahn2022saycan,shi2025hi}. We use a skill query to condition a
representation of the observed interaction. Short observation windows in
policy learning~\citep{chi2023diffusion} and longer sequence context
\citep{chen2021decision} offer different temporal information. Our sparse memory
retains skill-terminal observations and is evaluated against comparable recent
and uniformly sampled histories \cite{zeng2026kemo,yang2026eventvla,shi2026memoryvla}. 


\paragraph{Progress supervision and representation evaluation.}
VIP and LIV connect learned visual representations to reward and task structure
\citep{ma2022vip,ma2023liv}; ReWiND and Robometer study learned rewards and their
use in robot learning~\citep{rewind2025,robometer2025,chen2026sarm}. We use progress supervision
to train and diagnose a multimodal representation, concentrating on contact
states and historical dependencies. Comparisons with existing progress models
measure how well the representation supports progress estimation. We separately
evaluate the representation itself by freezing its features and training simple
probes on top, testing whether it captures task-relevant information beyond what
is needed by the original progress predictor.

\section{Structuring Skill-conditioned Representations with Event Memory}
\label{sec:method}
We introduce \AlgName{}, a skill-conditioned Transformer that combines
recent visual and tactile observations with a sparse memory of earlier
skill events (Figure~\ref{fig:model}). We first define primitive skills,
the skill-conditioned representation, and skill progress targets
(Section~\ref{sec:formulation}). We then describe how visual and tactile
observations are encoded over modality-specific histories
(Section~\ref{sec:features}), how skill-conditioned attention combines
current observations with retained events
(Section~\ref{sec:attention}), and how observation windows are retained
at detected skill terminations to form a sparse event memory
(Section~\ref{sec:memory}). Finally, we describe how skill-progress
supervision trains the representation and how predicted progress
is used to construct memory during training
(Section~\ref{sec:learning}).

\subsection{Skill-conditioned formulation}
\label{sec:formulation}

Long-horizon manipulation comprises primitive skills with distinct
objectives and sensory requirements: reaching brings the gripper to
a target, guided predominantly by vision; tightening secures a part,
relying on tactile feedback; and alignment establishes the required
relative position and orientation, using both modalities.

\paragraph{Skills as the unit of long horizon tasks.}
Long-horizon manipulation comprises primitive skills with distinct
objectives and sensory requirements: reaching brings the gripper to
a target, guided predominantly by vision; tightening secures a part,
relying on tactile feedback; and alignment establishes the required
relative position and orientation, using both modalities. This
decomposition structures representation learning around local
objectives and enables skill-dependent attention to multi-modal
observations. The same observed state may satisfy one skill’s objective but not another’s. We therefore condition the representation on the queried skill when estimating progress and completion. Associating representations with
primitive objectives also provides a basis for reusing skills in
different orders and task compositions. To assess whether the representation captures these skill-specific distinctions, we use skill progress as an interpretable readout of intermediate states and completion. Its definition depends on how each interaction unfolds, motivating the following progress types.


\paragraph{Skill progress types.}
A primitive skill is a goal-directed interaction with a language-described
objective and an observable termination condition. We distinguish
continuous and discrete skills, with conditional skills extending the
discrete formulation to include failed outcomes.

\emph{Continuous} skills, such as reaching, have observable advancement
throughout execution. We assign dense progress targets by linearly
normalizing time within the annotated skill interval:
\[
p_t^k =
\frac{t - t_k^{\mathrm{start}}}
     {t_k^{\mathrm{end}} - t_k^{\mathrm{start}}},
\qquad
t \in [t_k^{\mathrm{start}}, t_k^{\mathrm{end}}].
\]

\emph{Discrete} skills, such as sliding a bottle cap or tightening a bulb,
have intermediate observations that do not reliably indicate progress.
Completion is instead marked by a terminal event. Progress remains at
0 during execution and rises sharply to 1 over the last few frames
preceding the annotated termination.

\emph{Conditional} skills, such as inspection, are discrete skills that
can terminate in success or failure. For these skills, we use
three-valued targets: $p_t^k = 0$ before termination, $p_t^k = 1$ at
successful termination, and $p_t^k = -1$ at failed termination.
The terminal outcome determines the subsequent sequence of skills.

\begin{figure}[t]
  \centering
   \begin{tikzpicture}
      \node[anchor=south west, inner sep=0] (img)
        {\includegraphics[width=0.9\linewidth]{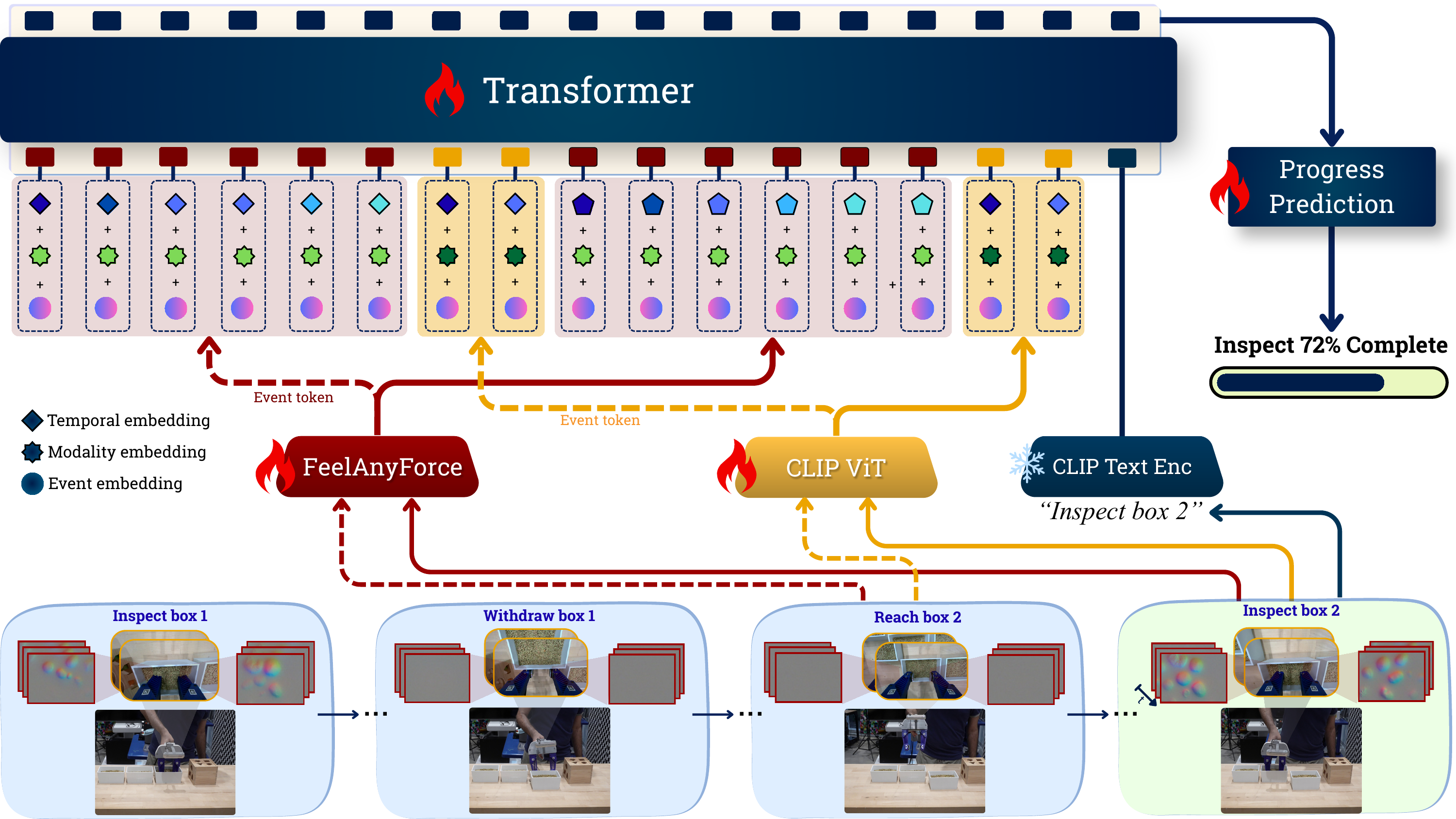}};
       
    \end{tikzpicture}

  \caption{\AlgName{} on \emph{blind cube search}, queried with
  \emph{``inspect box 2''}. \textbf{What to attend:} the two tactile streams
  are encoded frame-by-frame by FeelAnyForce and the RGB stream by CLIP ViT
  (both fine-tuned), each over its own horizon; the skill description is
  encoded by the frozen CLIP text tower. Every token is the sum of
  its feature and a temporal, modality, and event embedding, so the
  transformer knows when a token was observed, from which sensor, and whether
  it belongs to the current window or to memory. \textbf{What to keep:} when a
  skill completes (here \emph{inspect box 1}, \emph{withdraw box 1}, \emph{Reach box 2}) their encoded observation window
  is stored as event tokens and attended alongside the current
  observation. The learned output tokens are mapped to predict queried skill  progress.}
  \label{fig:model}
\end{figure}

\paragraph{Skill-conditioned state representation.}
We use a predefined vocabulary of primitive skills and annotate
demonstrations across our three tasks as sequences of executed skill
instances. We use language to identify the queried primitive skill within a predefined vocabulary, rather than to support open-vocabulary generalization.  Each instance $\ell_k$ has annotated start and end times
$(t_k^{\mathrm{start}},t_k^{\mathrm{end}})$, from which we derive progress
targets to supervise representation learning. At timestep $t$, the robot
receives an RGB image and two tactile images,
$o_t=(v_t,u_t^L,u_t^R)$. Conditioned on the queried skill $\ell_k$, the
model combines recent multi-modal observation windows $O_t$ with event
memory $M_t$, which retains observation windows from up to $K$ preceding
skill terminations:
\begin{equation}
 h_t^k=F_\theta(O_t,M_t,\ell_k)\in\mathbb{R}^{d}.
 \label{eq:representation}
\end{equation}
Here $O_t=(V_t,T_t)$ contains recent RGB and dual-tactile observations,
while $M_t$ retains observation windows from earlier skill events.
In the blind-search example in Figure~\ref{fig:model}, the query is
\emph{``inspect box 2''}. The green panel shows the current observation
windows $O_t$, and the preceding blue panels illustrate the events
retained in $M_t$. Current and retained observations follow the solid
and dashed paths, respectively, through the sensory encoders.
Their tokens are then processed together with the skill query
to estimate progress for the inspection of box 2.

Task-conditioned models such as ReWiND~\citep{rewind2025} and
Robometer~\citep{robometer2025} estimate progress relative to a task-level
instruction. Our formulation makes each primitive's objective explicit
during observation fusion, allowing the representation to weight
visual and tactile observations according to the queried skill.
Skill-level progress supervision also provides a mechanism for
identifying terminal observations to retain, connecting what the
representation attends to with what it keeps for subsequent interactions.

\subsection{Modality-specific features and histories}
\label{sec:features}

We encode visual and tactile observations separately using
CLIP ViT-B/16~\citep{radford2021learning} and
FAF~\citep{Shahidzadeh2024FeelAnyForceEC}, respectively.
For each frame, we extract a single CLS token from the corresponding
encoder and project it to $d=512$ dimensions. The tactile projection
includes layer normalization and is shared between the two tactile
sensors. These tokens are passed individually to the fusion network,
allowing it to attend to each sensor's features. We use observation
windows of two RGB frames and six frames from each tactile sensor,
reflecting the different temporal characteristics of vision and tactile sensing \cite{feng2025anytouch}.

For modality $m$, a window contains $H_m$ observations sampled with stride
$d_m$. Its temporal span is $(H_m-1)d_m/\nu_m$ seconds at sensor rate $\nu_m$.
Each observation is encoded as
\begin{equation}
 z^m_{t,j}=P_m f_m(o^m_{t-(H_m-1-j)d_m})+
 \eta_j^{\rm time}+\eta_m^{\rm modality}+\eta_0^{\rm segment},
 \label{eq:tokens}
\end{equation}

  where $j=0,\ldots,H_m-1$ indexes observations from oldest to newest,
  $f_m$ is the sensory encoder, and $P_m$ is the feature projection. 
  The learned embeddings encode temporal position, sensor identity. Retained tokens
  additionally receive event-slot embeddings indicating chronological
  order. Horizons and encoders are shared across skill queries;
  the fusion transformer learns how to utilize their tokens for each skill.

\subsection{Skill-conditioned Attention}
\label{sec:attention}
The transformer in Figure~\ref{fig:model} jointly processes the skill
query, current observation tokens, and retained event tokens.
Self-attention allows interactions across sensors and observation
windows, while a learned state token aggregates their information
into a representation for the queried skill.

We encode the skill description with the frozen CLIP text encoder and
project it to a query token $q_k$. Given current observation tokens $Z_t$,
encoded event memory $E_\theta(M_t)$, and a learned state token
$z_{\mathrm{out}}$, the representation is
\begin{equation}
 h_t^k = \operatorname{LN}\!\left(
 \operatorname{Transformer}_\theta
 \bigl([q_k; Z_t; E_\theta(M_t); z_{\mathrm{out}}]\bigr)_{\mathrm{out}}
 \right).
 \label{eq:fusion}
\end{equation}

Attention is bidirectional within the input sequence, which includes
recent observation tokens and tokens from earlier event windows
retained in $M_t$. The state token can therefore combine recent
observations with past interactions relevant to the queried skill.
No observations after time $t$ are used.

\subsection{Long- and Short-Term Memory via Sparse Events}
\label{sec:memory}
The recent observation window $O_t$ provides short-term context, while
event memory $M_t$ retains earlier observations that may remain useful
for subsequent skills. When a skill event is detected at frame $t$,
we add the sensor window ending at that frame, $O_t^{\rm event}$,
to memory:
\begin{equation}
M_{t+1}=\operatorname{LastK}\big(M_t\mathbin{\|}
O_t^{\rm event}\big).
\label{eq:memory}
\end{equation}
Here $\|$ appends the new window, and $\operatorname{LastK}$ retains
at most $K$ of the most recent event windows in chronological order.
Memory remains unchanged when no event is detected.

Event detection at time $t$ uses the current observations $O_t$ and
existing memory $M_t$. The newly stored window becomes available at
$t+1$, so it cannot affect its own detection. We use $K=1$ for lamp and bottle and $K=3$ for blind search.
In Figure~\ref{fig:model}, the three retained events correspond to
\emph{inspect box 1}, \emph{withdraw box 1}, and \emph{reach box 2}.
Memory preserves the observation windows ending at these events,
rather than the full sequence of intervening observations.

We detect events using the progress predictor described in
Section~\ref{sec:learning}. For the skill instance $a_t$ queried during
event construction, a single frame satisfying
$|\hat p_t^{a_t}|\geq\tau_{a_t}$ is sufficient to add the observation
window ending at that frame to memory. We use a threshold of $0.95$
for continuous skills and a lower threshold of $0.25$ for discrete
and conditional skills, whose targets remain zero until near
termination and then change sharply. Taking the absolute value
allows both successful and failed conditional outcomes to trigger
an event.

\subsection{Training the skill progress predictor}
\label{sec:learning}
The progress-prediction block in Figure~\ref{fig:model} applies a head
$g_\psi$ to the skill-conditioned representation $h_t^k$, producing
$\hat p_t^k=g_\psi(h_t^k)$. We train the model by minimizing the
squared error between predicted progress and the annotated targets:
\begin{equation}
 \mathcal{L}(\theta,\psi)=\mathbb{E}_{(t,k)}
 [(g_\psi(h_t^k)-p_t^k)^2].
 \label{eq:loss}
\end{equation}
With probability $\alpha$, we query the active skill; otherwise,
we sample uniformly from the inactive skill descriptions. Pairing the same observations with different skill queries
encourages the model to estimate \textbf{progress for the queried skill},
rather than always for the skill currently being performed.
This encourages attention to visual and tactile features
\textbf{relevant to the queried skill}. Visual and tactile encoders are fine-tuned with the fusion transformer, while the text encoder remains frozen. 

Because event selection relies on predicted progress, we first train
the model without past event observations, then introduce event memory.

\paragraph{Warmup training without event memory.}
We first train the model to predict skill progress from the current
observation windows. Event slots remain in the input sequence, but
their sensory inputs are set to zero. After this initial training,
we run the predictor over each training episode and identify event
frames using the progress thresholds in Section~\ref{sec:memory}.
The observation windows ending at these frames are retained for
the next stage.

\paragraph{Training with event memory.}
We continue training the same model using both current observations
and the selected event windows. Every $R$ epochs, we run the updated
predictor over all training episodes to select a new set of event
frames. Between these updates, the selected frames remain fixed,
while the model continues to learn from current and retained
observations. Progress targets are still derived from the annotations;
the model's predictions determine which past observation windows
are retained in memory.



\section{Contact-rich Datasets and Tasks}
\label{sec:datasets}
We study three contact-rich manipulation tasks with complementary sensing
and memory requirements (Figure S\ref{fig:tasks}). \textbf{Lamp replacement}
combines reaching and grasping with thread alignment and tightening.
Successive twists look similar, while touch captures increased resistance
at completion; we disable the light-on cue to remove this visual shortcut.
\textbf{Bottle capping} requires detecting a brief slip during repetitive
cap sliding to transition to the final twist. Both tasks involve thread
alignment, but their distinct contact patterns require object-specific
interpretation of tactile observations. \textbf{Blind cube search} involves searching among three boxes containing a
cube or distractors (a cylinder and a pentagonal prism), withdrawing after
unsuccessful inspections, and inserting the cube into a puzzle box once
found. Touch helps identify the shape, while inspection history and
outcomes determine progress for box-specific skills. For example, finding
the cube in box 1 leaves ``reach box 2'' unexecuted, so its progress should
remain zero. During later insertion, this history also identifies which
box contained the cube. Event memory retains context for these distinctions
beyond the current observation window. Together, these tasks test short-term contact dynamics and longer-term context
retained through sparse events. They provide a
setting to evaluate whether sparse event memory preserves useful task
history beyond short-term observation windows without retaining a dense
sequence of past observations.
See the dataset section in the supplementary material for task details,
skill definitions, and termination conditions.
\begin{wrapfigure}{r}{0.2\textwidth}
\centering
\vspace{-0.4cm}
\includegraphics[width=\linewidth]{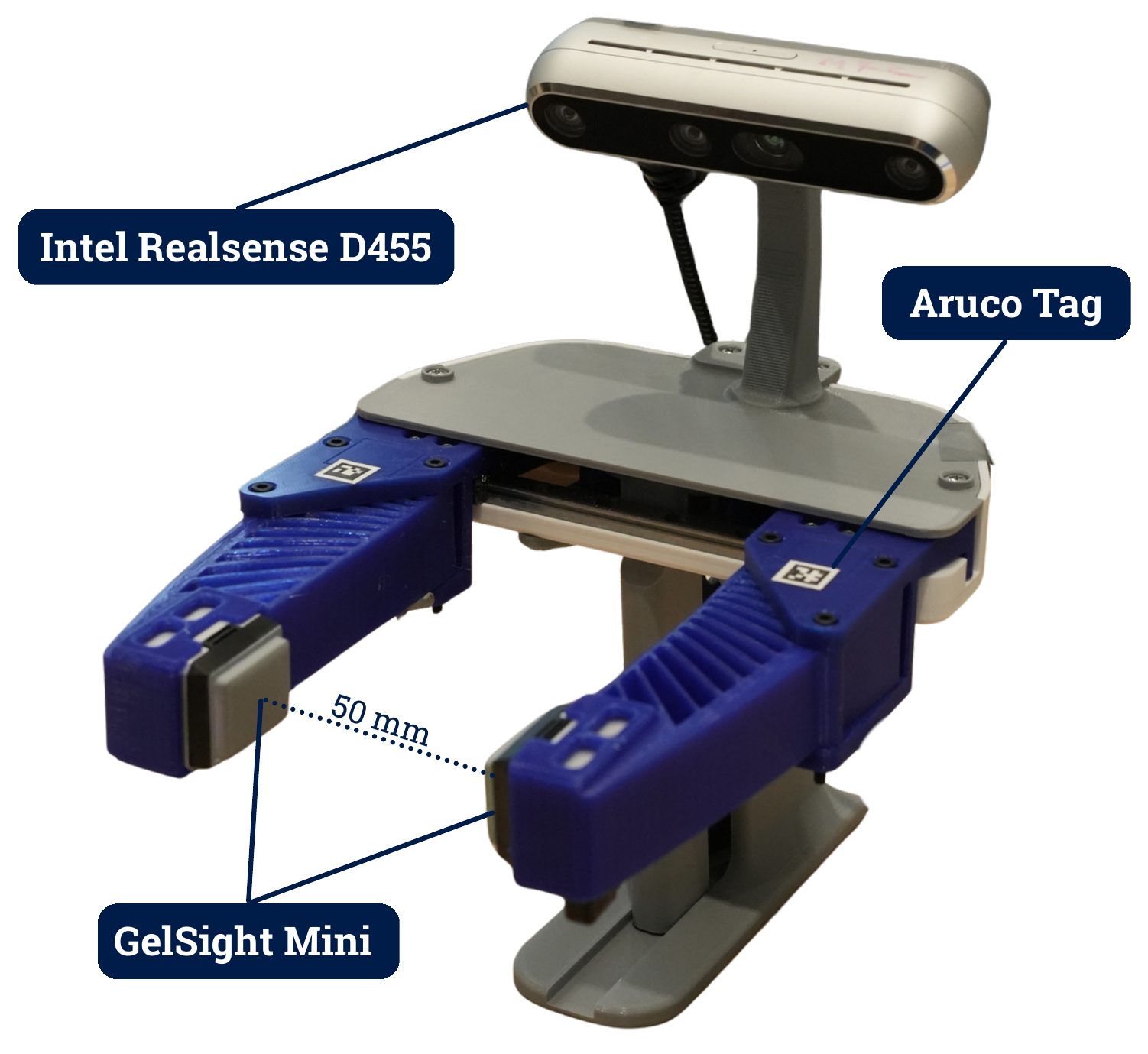}
\vspace{-0.5cm}
\caption{\footnotesize\textbf{UMI-Tac.} a handheld gripper with a wrist camera and tactile sensors.}
\vspace{-1cm}
\label{fig:umitac}
\end{wrapfigure}
\paragraph{Hardware setup.}
Demonstrations are collected with UMI-Tac, a customized UMI gripper
\citep{chi2024universalmanipulationinterfaceinthewild}, equipped with two
GelSight Mini tactile sensors at fingertips and an Intel RealSense D455 RGB-D wrist camera
(Figure~\ref{fig:umitac}). Synchronized RGB and dual-tactile image streams
provide visual context and contact observations for all three tasks.


\paragraph{Dataset size and split.}
The datasets contain 429 episodes in total: 134 for lamp replacement, 151
for bottle capping, and 144 for blind cube search. Approximately 10\% of each
original dataset is held out at the episode level as a test set.

\section{SkillFormer Remembers Task-Relevant Events}
\label{sec:exp}
\vspace{-0.1cm}
In this section, we compare \AlgName{} to a number of language-conditioned representation baselines. Across our challenging tasks, \AlgName{} effectively prioritizes and remembers the task-relevant events,  reducing event detection delays and improving progress estimation. We describe baselines and metrics and expose the core design decisions that contribute to the strong performance relative to other multi-modal or vision-only alternatives that do not employ memory.  

\paragraph{Baselines.}
  We compare \AlgName{} against two language-conditioned manipulation reward models and four ablations. Robometer is a vision-language reward model built on Qwen3-VL-4B
  and pretrained on large-scale datasets with both per-frame progress and
  cross-trajectory preference objectives. We report it zero-shot and after fine-tuning a rank-32 LoRA
  adapter and its progress heads on our dataset noted as Robometer$^*$. ReWiND employs a transformer  over frozen DINOv2 image and MiniLM language features; we train it on our dataset together with Open X-Embodiment \cite{o2024open} data noted as ReWiND$^*$. Both use RGB observations without touch or event memory. Our ablations isolate individual design
  choices under the matched data and evaluation protocol: \textbf{Vision Only} removes the
  tactile encoders; \textbf{\woevent{}} removes the event tokens; \textbf{\film{}} replaces our bidirectional self-attention with FiLM conditioning of the visual
  features on tactile features and task description; and \textbf{\AlgName{} ($h{=}2$)}
  shortens the tactile horizon from six to two frames. \textbf{w/o FAF} replaces FAF with an ImageNet-pretrained DINOv2. On blind search, we additionally evaluate \AlgName{} with a single event slot ($K{=}1$) instead of three, and the vision-only ablation is trained with $K{=}3$ to match the full model.

\begin{table}[t]
\centering
\caption{Progress estimation results across various visual or tactile dominant skills.{\scriptsize(\textbf{bold} and \underline{underline} are based on the best mean)}}
\label{tab:baselines_short}
\small
\setlength{\tabcolsep}{3pt}
\renewcommand{\arraystretch}{0.4}
\resizebox{\columnwidth}{!}{
\begin{tabular}{l l cccc}
\toprule
\multirow{2}{*}{\textbf{Method}} & \multirow{2}{*}{\textbf{Metric}} & \multicolumn{2}{c}{\textbf{Lamp replacement}} & \multicolumn{2}{c}{\textbf{Bottle capping}} \\
\cmidrule(lr){3-4}\cmidrule(lr){5-6}
& & Reach bulb & Twist bulb & Reach cap & Slide cap \\
\midrule
\multirow{2}{*}{Robometer~\cite{robometer2025}}
& Pearson $r$ $\uparrow$ & $0.638 \pm 0.054$ & $0.209 \pm 0.053$ & $0.819 \pm 0.084$ & $0.578 \pm 0.058$ \\
& MSE ($\times10^{-3}$) $\downarrow$ & $84.5 \pm 19.6$ & $499.8 \pm 59.3$ & $46.8 \pm 17.4$ & $232.2 \pm 42.2$ \\
\midrule
\multirow{2}{*}{Robometer$^*$~\cite{robometer2025}}
& Pearson $r$ $\uparrow$ & $0.886 \pm 0.032$ & $0.370 \pm 0.106$ & \underline{$0.885 \pm 0.018$} & $0.875 \pm 0.020$ \\
& MSE ($\times10^{-3}$) $\downarrow$ & $27.4 \pm 6.2$ & $98.6 \pm 46.3$ & $51.3 \pm 9.5$ & \underline{$40.7 \pm 4.3$} \\
\midrule
\multirow{2}{*}{ReWiND$^*$~\cite{rewind2025}}
& Pearson $r$ $\uparrow$ & $0.847 \pm 0.060$ & $0.635 \pm 0.136$ & $0.880 \pm 0.035$ & \underline{$0.878 \pm 0.037$} \\
& MSE ($\times10^{-3}$) $\downarrow$ & \underline{$13.7 \pm 10.4$} & $21.4 \pm 10.2$ & $11.5 \pm 4.7$ & $45.5 \pm 11.9$ \\
\midrule
\multirow{2}{*}{Vision Only}
& Pearson $r$ $\uparrow$ & \underline{$0.997 \pm 0.002$} & \underline{$0.842 \pm 0.095$} & \boldmath$0.997 \pm 0.003$\unboldmath & $0.798 \pm 0.103$ \\
& MSE ($\times10^{-3}$) $\downarrow$ & \boldmath$0.2 \pm 0.1$\unboldmath & \underline{$9.3 \pm 6.4$} & \boldmath$0.3 \pm 0.3$\unboldmath & $72.7 \pm 45.5$ \\
\midrule
\multirow{2}{*}{\AlgName{}}
& Pearson $r$ $\uparrow$ & \boldmath$0.998 \pm 0.001$\unboldmath & \boldmath$0.899 \pm 0.061$\unboldmath & \boldmath$0.997 \pm 0.003$\unboldmath & \boldmath$0.948 \pm 0.073$\unboldmath \\
& MSE ($\times10^{-3}$) $\downarrow$ & \boldmath$0.2 \pm 0.1$\unboldmath & \boldmath$5.9 \pm 3.9$\unboldmath & \underline{$0.4 \pm 0.3$} & \boldmath$15.2 \pm 20.6$\unboldmath \\
\bottomrule
\end{tabular}
}
\vspace{-0.5cm}
\end{table}

\paragraph{Evaluation pipeline.}
As described in Section~\ref{sec:method}, we learn a multi-modal representation that detects skill transitions with low delay and
uses retained events to interpret subsequent observations. We query every skill throughout each held-out episode and compare predicted
progress with ground-truth targets derived from annotated skill boundaries and progress types. Progress accuracy measures how well
the representation distinguishes skill advancement and completion over the entire episode, while event-detection delay measures how accurately it detects the events for memory retention. We report \textbf{MSE} to measure progress prediction error, \textbf{Pearson $\mathbf{r}$} correlation to measure agreement with the target's temporal variation, and \textbf{Delay@0.9} to measure the absolute timing error between predicted and ground-truth crossings of the 0.9 progress threshold.

\paragraph{\AlgName{} excels at skills that require reading tactile patterns over time.}
 Table~\ref{tab:baselines_short} shows that vision suffices for reaching, 
  but struggles with repetitive contact interactions. On bulb tightening, 
  \robometer{}$^*$ and \rewind{}$^*$ achieve correlations of only $0.370$ 
  and $0.635$, compared with $0.899$ for our method, indicating difficulty 
  tracking fine-grained skill progress. Relative to Vision Only, 
  \AlgName{} reduces tightening and cap-sliding MSE by 37\% and 79\%, 
  supporting the value of tactile cues for distinguishing resistance 
  and slip when successive motions look similar.

\begin{wrapfigure}{r}{0.50\textwidth}
\raggedleft
\vspace{-0.5cm}
\includegraphics[width=0.5\textwidth]{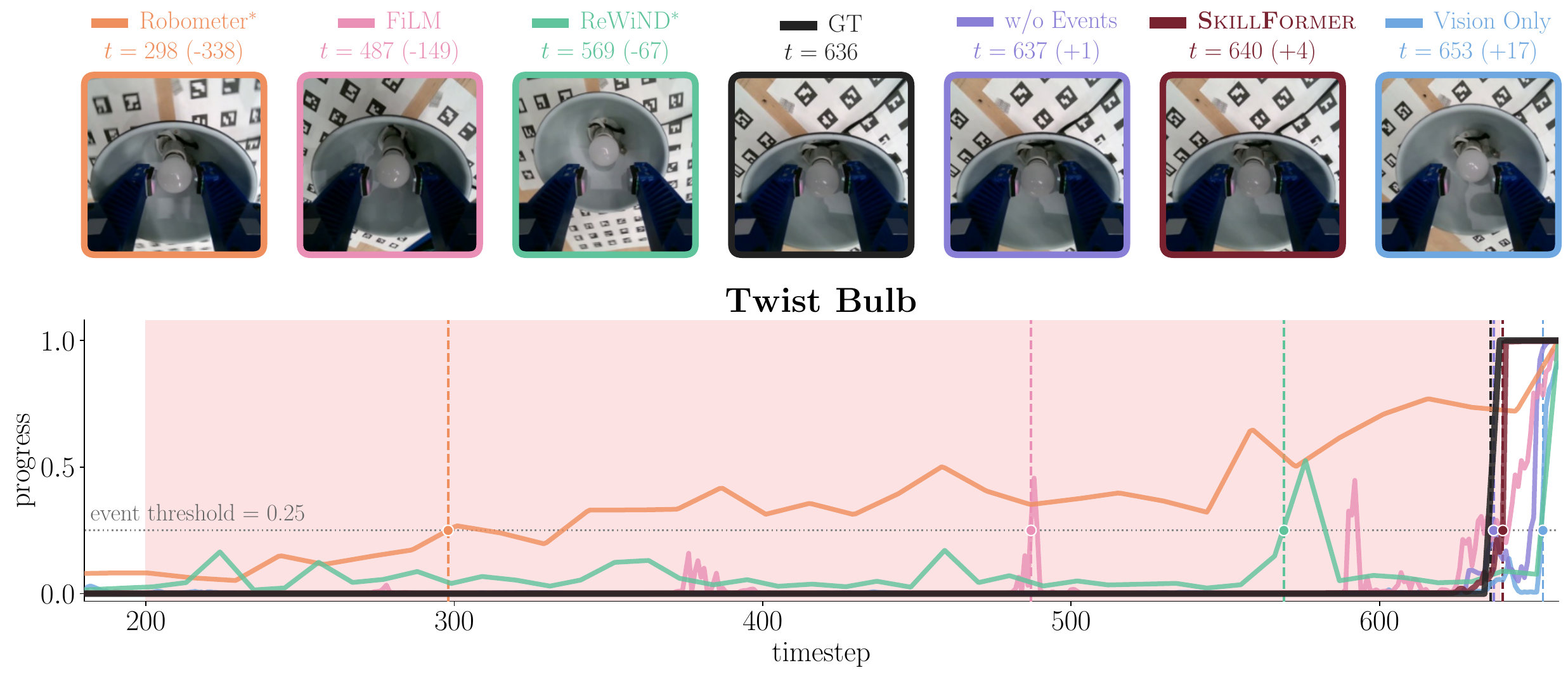}
\includegraphics[width=0.5\textwidth]{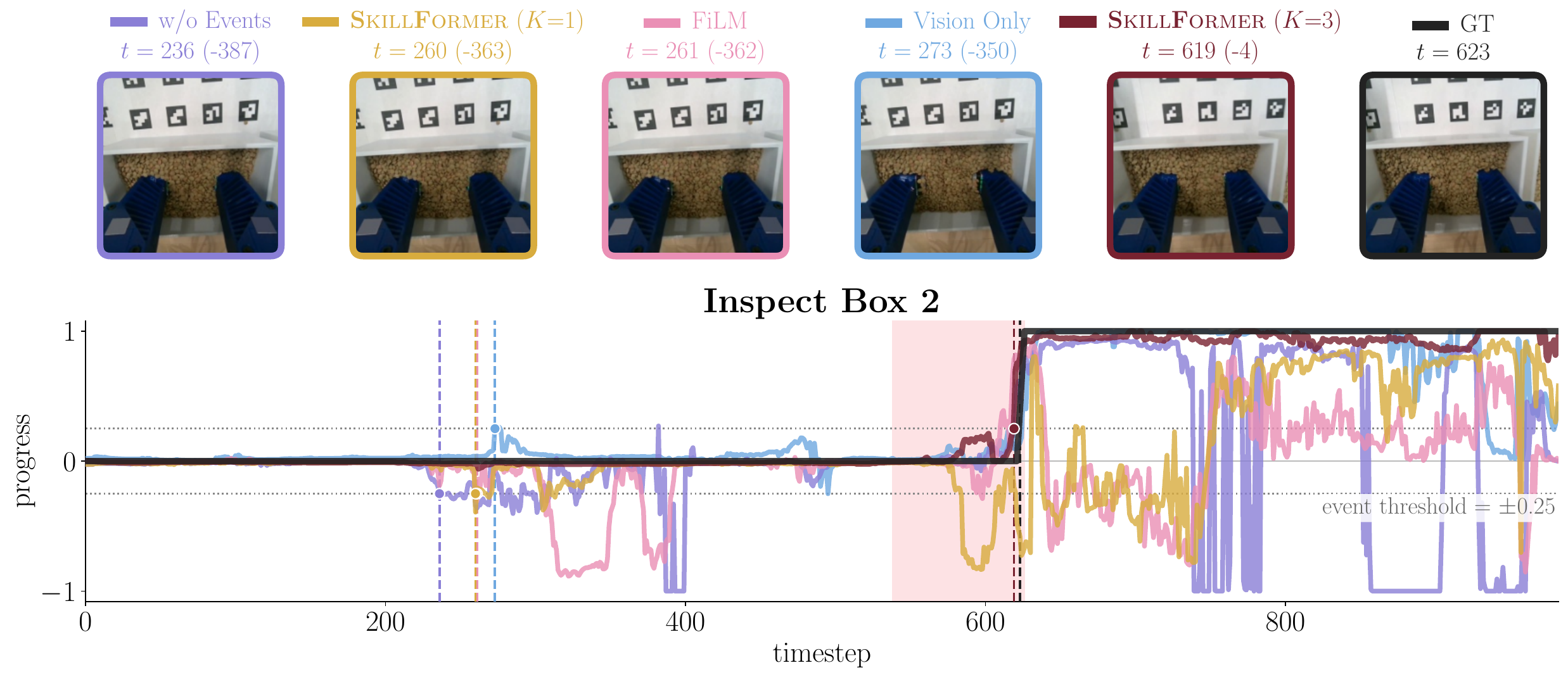}
\vspace{-0.8cm}
\caption{ We show observations and their timesteps which correspond to moments when an event is being detected where progress crosses the event threshold for a sample discrete \textbf{(top)} and conditional skill \textbf{(bottom)} in \textit{Lamp Replacement} and \textit{Blind Cube Search} tasks.}
\vspace{0.4 cm}
\label{fig:pivots}
\end{wrapfigure}

\begin{wrapfigure}{r}{0.5\columnwidth}
  \vspace{-0.9cm}
  \centering
  \begin{tikzpicture}
    \node[anchor=south west, inner sep=0] (imgA) at (0,0) {
      \includegraphics[width=0.50\columnwidth]{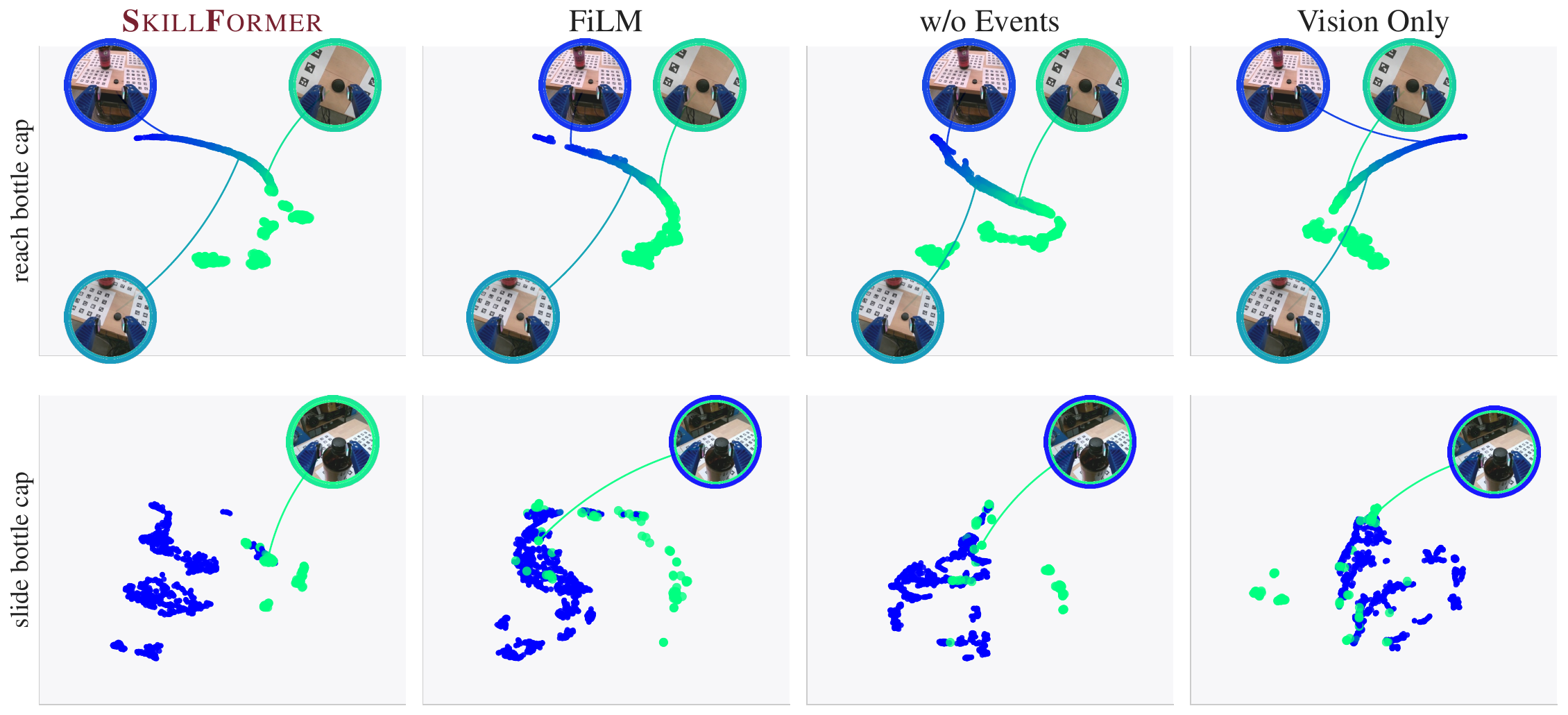}
    };
  \end{tikzpicture}\\[2pt]
  \vspace{-0.3cm}
  \begin{tikzpicture}
    \node[anchor=south west, inner sep=0] (imgB) at (0,0) {
      \includegraphics[width=0.50\columnwidth]{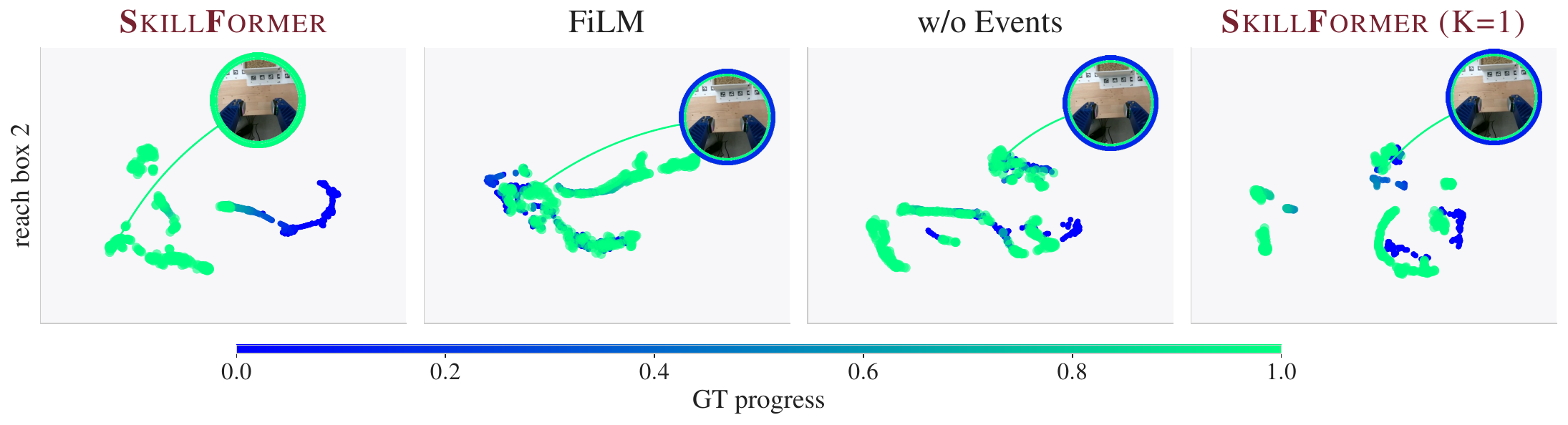}
    };
  \end{tikzpicture}
    \vspace{-0.7cm}
  \caption{\textbf{t-SNE projections} of produced transformer tokens colored by ground-truth progress. \small{Inner/outer image borders indicate gt/predicted progress.}}
  \label{fig:tsne}
  \vspace{-0.4cm}
\end{wrapfigure}

\paragraph{Event Memory resolves the representation ambiguity.}
In Figure~\ref{fig:pivots}, \robometer{}, \film{}, and \rewind{} trigger
before the bulb is fully tightened prematurely. The \woevent{} model detects the
event only one frame late, but its prediction subsequently drops,
effectively \emph{forgetting} completion. \AlgName{} triggers only
four frames late while visual observations still resemble ongoing
twisting, suggesting that it prioritizes tactile observations to differentiate between
the last twist and earlier twists without a tightening tactile pattern. For \inspecttwo{}, \woevent{} and \kone{} predict a negative outcome during an earlier \inspectone{}, when the \inspecttwo{} progress remains zero. This suggests confusion between similar
inspections across boxes. In contrast, \kthree{} detects the positive
outcome a few frames before the ground-truth crossing and 
preserves it afterward, supporting the value of retained history for
distinguishing and remembering inspection outcomes. Tab. \ref{tab:blind_short} further illustrates this in the blind cube search task where a vision-only transformer with 3 stored events outperforms our \kone{} variant. Moreover, the feature projections in 
  Figure~\ref{fig:tsne} provide complementary qualitative evidence: 
  for the same observations queried with \textit{Reach Box 2}, 
  \AlgName{} produces clearer separation between progress states, 
  while the other configurations exhibit greater overlap. For a similar continuous \textit{Reach Bottle Cap}, all models show a smooth progression from early reaching states to completion, consistent with the current visual observations providing sufficient context for this skill.

\paragraph{Limitations of naïve multi-modal token fusion.} 
  Tactile VLAs such as ~\cite{cheng2026omnivtlavisiontactilelanguageactionmodelssemanticaligned} and 
  Tactile-VLA~\cite{huang2025tactilevlaunlockingvisionlanguageactionmodels} concatenate visual, tactile, 
  and language tokens for joint transformer processing. Our \woevent{} 
  ablation follows this fusion pattern over recent observations, 
  while \film{} uses tactile-conditioned feature modulation. 
  The \htwo{} ablation uses two timesteps for both modalities, 
  testing the benefit of extending the tactile horizon while keeping 
  the visual horizon fixed. In Table~\ref{tab:main_results_bottle_lamp}, 
  \film{} and \woevent{} exceed 41 frames of cap-sliding detection 
  delay, compared with 11.73 for \AlgName{}. Removing events also 
  more than doubles overall MSE on both tasks, showing the importance 
  of retaining event observations that can never be recovered later. Restricting 
  both modalities to the same two timesteps increases cap-sliding delay to 
  25.55 frames and bulb twist to 20.07, suggesting that tactile observations benefit from a longer temporal context than vision for skills that require detecting a tactile pattern over a longer horizon. Replacing 
  FAF with generic visual pretraining further weakens the benefit of 
  touch: align cap MSE rises from $6.4$ to $12.9$, approaching 
  Vision Only's $18.4$. Tactile inputs therefore 
  benefit from both appropriate temporal context and specialized 
  features.


\begin{wraptable}{r}{0.55\columnwidth}
\vspace{-0.8cm}
\centering
\caption{Blind search results averaged over all subtasks.}
\label{tab:blind_short}
\small
\setlength{\tabcolsep}{3pt}
\renewcommand{\arraystretch}{0.98}
\resizebox{0.53\columnwidth}{!}{
\begin{tabular}{l ccc}
\toprule
\textbf{Method} & Delay@0.9 $\downarrow$ & MSE ($\times10^{-3}$) $\downarrow$ & Pearson $r$ $\uparrow$ \\
\midrule
w/o Events          & $63.46 \pm 119.43$ & $143.6 \pm 220.0$ & $0.683 \pm 0.301$ \\
FiLM                & $77.81 \pm 118.55$ & $116.0 \pm 253.4$ & $0.727 \pm 0.263$ \\
\AlgName{} (K=1)    & $42.77 \pm 79.83$  & $83.1 \pm 201.7$  & $0.791 \pm 0.287$ \\
Vision Only (K=3)   & \underline{$37.46 \pm 87.46$} & \underline{$57.1 \pm 155.1$} & \underline{$0.857 \pm 0.213$} \\
\AlgName{} (K=3)    & \boldmath$16.10 \pm 37.20$\unboldmath & \boldmath$11.6 \pm 34.1$\unboldmath & \boldmath$0.953 \pm 0.082$\unboldmath \\
\bottomrule
\end{tabular}
}
\vspace{-0.3cm}
\end{wraptable}

\begin{figure*}[t]
\centering
\includegraphics[width=1\textwidth]{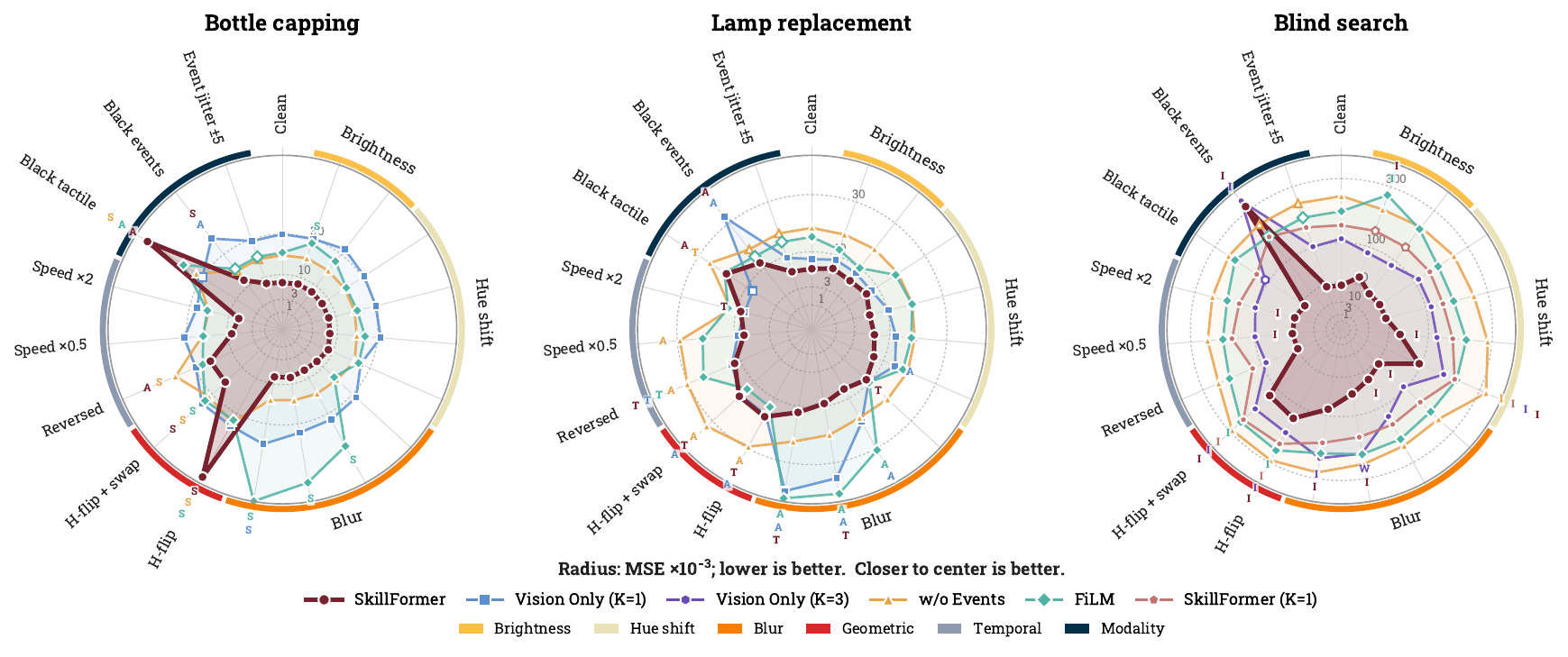}
\vspace{-0.8cm}
\caption{Robustness to visual, geometric, temporal, and modality 
  perturbations across three tasks. 
  closer to the center is better. Colored letters identify the subtask 
  with the largest error increase over clean inputs 
  (R: reach, A: align, S: slide, T: twist, I: inspect, W: withdraw).}
  \vspace{-0.6cm}
\label{fig:robustness}
\end{figure*}

\paragraph{\AlgName{} is robust to visual and temporal perturbations.} 
Fig.~\ref{fig:robustness} evaluates every model under different types of perturbations. On bottle capping and lamp replacement, \AlgName{}'s error stays close to its clean value across darkening, hue shifts and blur up to 15 px, while FiLM and Vision Only degrade markedly under strong blur, FiLM's bottle
  capping error roughly tripling at 7\,px. Temporal perturbations are equally benign: halving or doubling the timestep downsampling speed changes \AlgName{}'s error by at most a few units ($\times10^{-3}$), and jittering every detected   event by $\pm5$ frames leaves it essentially unchanged on all three tasks, showing that the event memory can tolerate a few frames latency around the exact event frame as it might last for a few frames or in the observation window. Blind search is the most
  sensitive task, with strong hue shifts and blur raising \AlgName{}'s error, yet it remains below every ablation
  at every level. \AlgName{} degrades sharply only when the inputs it relies on are removed, blacked-out tactile on the
  contact-rich bottle skills and a blacked-out event memory on blind search.  These targeted failures provide evidence that \AlgName{} actively exploits the modalities and memory tokens prioritized for each skill, rather than succeeding while ignoring them. We refer the reader to the appendix for further discussion of the robustness tests.

\begin{table*}[h]
\centering
\vspace{-0.5 cm}
\caption{Progress estimation results across various tasks and skills. Overall metrics are reported over the entire skills in each task.}
\label{tab:main_results_bottle_lamp}
\small
\setlength{\tabcolsep}{3pt}
\renewcommand{\arraystretch}{1.05}
\resizebox{\linewidth}{!}{
\begin{tabular}{l l ccc>{\columncolor{gray!25}}c:ccc>{\columncolor{gray!25}}c}
\hline
\multirow{2}{*}{\textbf{Method}} & \multirow{2}{*}{\textbf{Metric}} & \multicolumn{4}{c}{\textbf{Lamp replacement}} & \multicolumn{4}{c}{\textbf{Bottle capping}} \\
\cline{3-6}\cline{7-10}
& & Reach bulb & Align bulb & Twist bulb & Overall & Reach cap & Align cap & Slide cap & Overall \\
\hline
\multirow{3}{*}{Vision Only}
& Delay@0.9 & $3.75 \pm 2.09$ & $15.50 \pm 17.22$ & $19.75 \pm 8.69$ & $10.08 \pm 11.48$ & $5.36 \pm 3.31$ & $29.64 \pm 25.97$ & $82.18 \pm 82.08$ & $27.67 \pm 46.81$ \\
& MSE ($\times10^{-3}$) & \boldmath$0.2 \pm 0.1$\unboldmath & $13.8 \pm 18.8$ & $9.3 \pm 6.4$ & \underline{$5.4 \pm 9.7$} & \boldmath$0.3 \pm 0.3$\unboldmath & $18.4 \pm 14.8$ & $72.7 \pm 45.5$ & $17.8 \pm 32.7$ \\
& Pearson $r$ & \underline{$0.997 \pm 0.002$} & \underline{$0.967 \pm 0.039$} & \underline{$0.842 \pm 0.095$} & $0.904 \pm 0.160$ & \underline{$0.997 \pm 0.003$} & $0.963 \pm 0.028$ & $0.798 \pm 0.103$ & $0.950 \pm 0.085$ \\
\hline
\multirow{3}{*}{w/o Events}
& Delay@0.9 & $4.17 \pm 6.24$ & $15.50 \pm 15.96$ & \underline{$9.92 \pm 9.30$} & \underline{$7.92 \pm 10.41$} & \underline{$5.00 \pm 3.41$} & $17.18 \pm 8.63$ & $41.09 \pm 14.99$ & $17.36 \pm 15.50$ \\
& MSE ($\times10^{-3}$) & \underline{$0.4 \pm 0.8$} & $40.3 \pm 53.0$ & \underline{$8.4 \pm 11.6$} & $9.0 \pm 26.4$ & \boldmath$0.3 \pm 0.2$\unboldmath & $8.1 \pm 7.7$ & $46.8 \pm 13.1$ & $11.2 \pm 18.1$ \\
& Pearson $r$ & $0.991 \pm 0.023$ & $0.915 \pm 0.095$ & $0.811 \pm 0.285$ & $0.906 \pm 0.173$ & \boldmath$0.998 \pm 0.002$\unboldmath & \underline{$0.984 \pm 0.015$} & $0.847 \pm 0.046$ & $0.961 \pm 0.061$ \\
\hline
\multirow{3}{*}{w/o FAF}
& Delay@0.9 & -- & -- & -- & -- & $6.00 \pm 4.24$ & $18.82 \pm 12.88$ & $48.27 \pm 12.13$ & $19.52 \pm 20.51$ \\
& MSE ($\times10^{-3}$) & -- & -- & -- & -- & \boldmath$0.3 \pm 0.3$\unboldmath & $12.9 \pm 11.3$ & $55.3 \pm 16.5$ & $13.6 \pm 21.2$ \\
& Pearson $r$ & -- & -- & -- & -- & \underline{$0.997 \pm 0.003$} & $0.974 \pm 0.023$ & $0.817 \pm 0.057$ & $0.953 \pm 0.071$ \\
\hline
\multirow{3}{*}{FiLM}
& Delay@0.9 & $5.08 \pm 2.63$ & $18.17 \pm 22.05$ & $11.75 \pm 18.66$ & $8.97 \pm 14.55$ & $7.73 \pm 5.71$ & \underline{$7.27 \pm 3.82$} & $41.82 \pm 15.10$ & \underline{$14.27 \pm 15.70$} \\
& MSE ($\times10^{-3}$) & \underline{$0.4 \pm 0.4$} & $26.1 \pm 24.7$ & $14.2 \pm 13.2$ & $14.7 \pm 22.1$ & $0.6 \pm 0.5$ & \underline{$8.0 \pm 8.6$} & $50.7 \pm 20.3$ & $11.6 \pm 20.6$ \\
& Pearson $r$ & $0.990 \pm 0.011$ & $0.939 \pm 0.051$ & $0.749 \pm 0.242$ & $0.828 \pm 0.276$ & $0.995 \pm 0.004$ & \underline{$0.984 \pm 0.016$} & $0.834 \pm 0.067$ & $0.960 \pm 0.069$ \\
\hline
\multirow{3}{*}{\AlgName{} (h=2)}
& Delay@0.9 & \underline{$2.92 \pm 2.66$} & \boldmath$11.83 \pm 10.12$\unboldmath & $75.75 \pm 170.18$ & $20.07 \pm 81.29$ & \boldmath$4.55 \pm 3.11$\unboldmath & $10.64 \pm 5.65$ & \underline{$25.55 \pm 29.64$} & $14.44 \pm 18.63$ \\
& MSE ($\times10^{-3}$) & \boldmath$0.2 \pm 0.1$\unboldmath & \boldmath$12.2 \pm 12.2$\unboldmath & $14.0 \pm 22.6$ & $6.2 \pm 13.0$ & \underline{$0.4 \pm 0.3$} & $11.2 \pm 12.9$ & \underline{$24.0 \pm 38.6$} & \underline{$7.7 \pm 18.9$} \\
& Pearson $r$ & \underline{$0.997 \pm 0.002$} & \boldmath$0.970 \pm 0.029$\unboldmath & $0.827 \pm 0.210$ & \underline{$0.914 \pm 0.176$} & $0.996 \pm 0.003$ & $0.978 \pm 0.025$ & \underline{$0.914 \pm 0.146$} & \underline{$0.973 \pm 0.068$} \\
\hline
\multirow{3}{*}{\AlgName{}}
& Delay@0.9 & \boldmath$2.33 \pm 1.31$\unboldmath & \underline{$13.92 \pm 14.92$} & \boldmath$6.42 \pm 3.25$\unboldmath & \boldmath$7.60 \pm 9.20$\unboldmath & $5.36 \pm 3.11$ & \boldmath$6.27 \pm 5.64$\unboldmath & \boldmath$11.73 \pm 16.58$\unboldmath & \boldmath$10.70 \pm 15.78$\unboldmath \\
& MSE ($\times10^{-3}$) & \boldmath$0.2 \pm 0.1$\unboldmath & \underline{$12.5 \pm 14.6$} & \boldmath$5.9 \pm 3.9$\unboldmath & \boldmath$3.4 \pm 7.7$\unboldmath & \underline{$0.4 \pm 0.3$} & \boldmath$6.4 \pm 5.8$\unboldmath & \boldmath$15.2 \pm 20.6$\unboldmath & \boldmath$4.6 \pm 10.3$\unboldmath \\
& Pearson $r$ & \boldmath$0.998 \pm 0.001$\unboldmath & \boldmath$0.970 \pm 0.033$\unboldmath & \boldmath$0.899 \pm 0.061$\unboldmath & \boldmath$0.971 \pm 0.045$\unboldmath & \underline{$0.997 \pm 0.003$} & \boldmath$0.987 \pm 0.012$\unboldmath & \boldmath$0.948 \pm 0.073$\unboldmath & \boldmath$0.985 \pm 0.035$\unboldmath \\
\hline
\end{tabular}
}
\end{table*}


\paragraph{Conclusion} 
We presented \AlgName{}, a skill-conditioned multimodal representation that combines tactile temporal context with event memory. Through extensive evaluation of common multimodal representations across three manipulation tasks, we demonstrate improved progress estimation and event detection that would not possible with simple multimodal concatenation or without event memory. Predicted progress could provide rewards for policy learning, while the representation could condition action generation on current observations and retained history. Evaluating these applications in closed-loop control remains future work.
  
\newpage

\bibliography{references}
\bibliographystyle{arxiv_references}
\newpage
\appendix

\section{Appendix}

\subsection{Tasks}
\label{sec:supp_tasks}

Each task is represented as an ordered sequence of language-addressable primitive skills. For example, bottle capping is decomposed into \emph{reach bottle cap}, \emph{grasp cap}, \emph{reach bottle}, \emph{align cap}, and \emph{slide cap}. Since progress is estimated conditioned on the queried skill description, the same representation can be reused across different phases of a task while changing only the language query. This formulation separates skill-level progress estimation from the particular long-horizon task sequence, enabling primitive skills to serve as reusable units for composing longer manipulation behaviors. We describe the task designs below.

\subsubsection{Lamp Replacement}
\label{sec:supp_lamp}

\begin{figure}[H]
    \centering
    \includegraphics[width=0.9\linewidth]{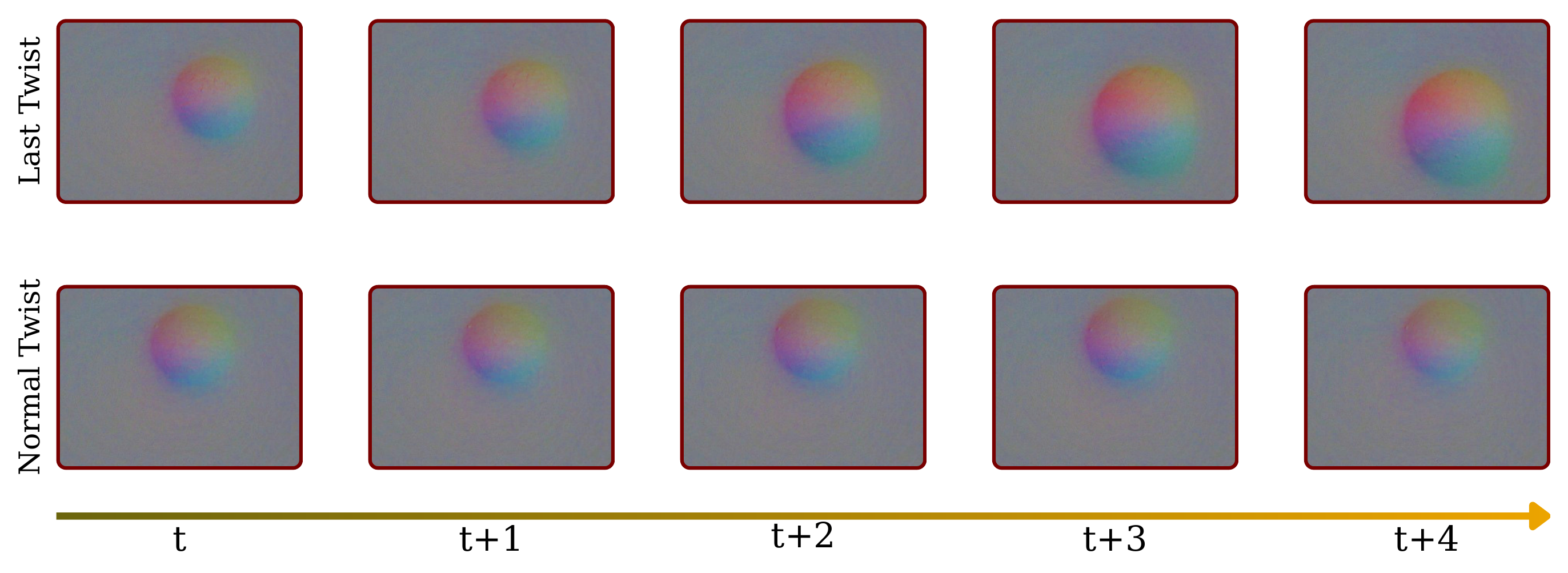}
    \caption{Tactile evolution during normal and final bulb twisting. During the last twist (top row), the tactile image shows an increasing-force pattern as resistance builds near the end of the thread. During normal twisting, the contact pattern remains relatively stable and changes only slightly across frames. This temporal tactile cue defines completion of the \emph{twist bulb} skill.}
    \label{fig:final_twist_pattern}
\end{figure}

In the lamp replacement task, the robot picks up a bulb, moves it toward the lamp socket, aligns it with the opening, and twists it until the bulb reaches the end of the thread. The task consists of five skills: \emph{reach bulb}, \emph{grasp bulb}, \emph{reach lamp socket}, \emph{align bulb}, and \emph{twist bulb}. The reaching skills are primarily visual, since progress is correlated with the relative position of the gripper and target object. In contrast, grasping and twisting require tactile feedback. The twist skill is especially contact-dependent: it repeats until a completion is indicated by increased tactile force (see Figure \ref{fig:final_twist_pattern}) once the bulb is fully tightened. This setting emphasizes the need for tactile temporal context; a single tactile frame may show contact, but the sequence of tactile frames reveals whether the interaction corresponds to normal rotation or terminal tightening. We disable the light-on visual cue so that twist completion cannot be inferred from appearance alone.

\subsubsection{Bottle Capping}

\label{sec:supp_bottle}

In the bottle capping task, the robot grasps a cap, moves it to the bottle opening, aligns it with the bottle, and slides it until terminal contact or slippage is detected. The task is decomposed into \emph{reach bottle cap}, \emph{grasp cap}, \emph{reach bottle}, \emph{align cap}, and \emph{slide cap}. This task tests whether the skill-conditioned representation can facilitate switching between vision-dominant and touch-dominant phases. Reaching and coarse alignment depend mostly on RGB observations, while grasping and sliding depend on tactile deformation. The slide skill is difficult because the cap can remain visually close to the bottle while the tactile signal changes sharply due to contact and slippage.

\begin{table*}[b]
\centering
\caption{Progress estimation results across blind cube search skills.}
\label{tab:main_results_blind}
\small
\setlength{\tabcolsep}{3pt}
\renewcommand{\arraystretch}{0.98}
\resizebox{\linewidth}{!}{
\begin{tabular}{l l cccc:c:}
\hline
\textbf{Method} & \textbf{Metric} & Reach box 1 & Inspect box 1 & Inspect box 2 & Withdraw 2 & Overall \\
\hline
\multirow{3}{*}{w/o Events}
& Delay@0.9 $\downarrow$ & $25.57 \pm 12.53$ & $31.00 \pm 15.43$ & $36.10 \pm 11.55$ & $66.20 \pm 96.58$ & $63.46 \pm 119.43$ \\
& MSE ($\times10^{-3}$) $\downarrow$ & $23.5 \pm 19.6$ & $496.0 \pm 455.7$ & $287.1 \pm 255.4$ & $131.4 \pm 89.4$ & $143.6 \pm 220.0$ \\
& Pearson $r$ $\uparrow$ & $0.833 \pm 0.128$ & $0.355 \pm 0.350$ & $0.462 \pm 0.449$ & $0.580 \pm 0.116$ & $0.683 \pm 0.301$ \\
\hline
\multirow{3}{*}{FiLM}
& Delay@0.9 $\downarrow$ & $58.14 \pm 25.27$ & $47.50 \pm 18.94$ & $100.80 \pm 46.24$ & \underline{$63.00 \pm 92.80$} & $77.81 \pm 118.55$ \\
& MSE ($\times10^{-3}$) $\downarrow$ & $14.7 \pm 13.2$ & $495.1 \pm 649.8$ & $171.7 \pm 132.2$ & $56.0 \pm 74.4$ & $116.0 \pm 253.4$ \\
& Pearson $r$ $\uparrow$ & $0.868 \pm 0.101$ & $0.400 \pm 0.316$ & $0.621 \pm 0.238$ & $0.575 \pm 0.250$ & $0.727 \pm 0.263$ \\
\hline
\multirow{3}{*}{\AlgName{} (K=1)}
& Delay@0.9 $\downarrow$ & $17.38 \pm 11.81$ & $61.67 \pm 67.35$ & $88.50 \pm 55.27$ & $198.60 \pm 177.28$ & $42.77 \pm 79.83$ \\
& MSE ($\times10^{-3}$) $\downarrow$ & $2.0 \pm 1.9$ & $388.6 \pm 513.1$ & $136.3 \pm 127.6$ & $47.2 \pm 81.6$ & $83.1 \pm 201.7$ \\
& Pearson $r$ $\uparrow$ & $0.986 \pm 0.012$ & $0.505 \pm 0.512$ & $0.705 \pm 0.274$ & $0.597 \pm 0.297$ & $0.791 \pm 0.287$ \\
\hline
\multirow{3}{*}{Vision Only (K=3)}
& Delay@0.9 $\downarrow$ & \boldmath$5.38 \pm 4.29$\unboldmath & \boldmath$7.33 \pm 6.99$\unboldmath & \underline{$15.00 \pm 7.10$} & $102.20 \pm 123.69$ & \underline{$37.46 \pm 87.46$} \\
& MSE ($\times10^{-3}$) $\downarrow$ & \underline{$1.3 \pm 1.6$} & \underline{$224.5 \pm 377.1$} & \underline{$131.0 \pm 194.3$} & \underline{$24.1 \pm 25.4$} & \underline{$57.1 \pm 155.1$} \\
& Pearson $r$ $\uparrow$ & \underline{$0.992 \pm 0.007$} & \underline{$0.712 \pm 0.343$} & \underline{$0.749 \pm 0.261$} & \underline{$0.900 \pm 0.039$} & \underline{$0.857 \pm 0.213$} \\
\hline
\multirow{3}{*}{\AlgName{} (K=3)}
& Delay@0.9 $\downarrow$ & \underline{$11.52 \pm 6.51$} & \underline{$8.33 \pm 4.99$} & \boldmath$8.70 \pm 8.04$\unboldmath & \boldmath$14.20 \pm 21.97$\unboldmath & \boldmath$16.10 \pm 37.20$\unboldmath \\
& MSE ($\times10^{-3}$) $\downarrow$ & \boldmath$0.9 \pm 0.7$\unboldmath & \boldmath$50.7 \pm 62.2$\unboldmath & \boldmath$18.6 \pm 24.3$\unboldmath & \boldmath$4.6 \pm 10.8$\unboldmath & \boldmath$11.6 \pm 34.1$\unboldmath \\
& Pearson $r$ $\uparrow$ & \boldmath$0.994 \pm 0.005$\unboldmath & \boldmath$0.898 \pm 0.106$\unboldmath & \boldmath$0.958 \pm 0.045$\unboldmath & \boldmath$0.986 \pm 0.023$\unboldmath & \boldmath$0.953 \pm 0.082$\unboldmath \\
\hline
\end{tabular}
}
\end{table*}

\subsubsection{Blind Cube Search}
\label{sec:supp_blind}

Blind cube search is designed to test tactile object identification. The robot searches candidate boxes containing hidden shapes, including a cube and distractor objects such as pentagonal prism and cylinder. Since the hidden object is not visible, the robot must infer whether it has found the cube from tactile deformation patterns. The task is decomposed into \emph{reach box}, \emph{inspect box}, \emph{withdraw}, \emph{reach puzzle box}, and \emph{insert cube}. If the inspected object is not the cube, the robot withdraws and reaches to the next candidate box. If the cube is detected, the robot carries it to the puzzle box and inserts it into the square hole. This task evaluates whether the representation can use contact geometry when visual observations are ambiguous or uninformative. Table \ref{tab:main_results_blind} provides a per-skill breakdown of the blind cube search results reported in Table \ref{tab:blind_short}. 
\begin{table}[H]
\renewcommand{\arraystretch}{0.9}
\centering
\caption{Skill definitions used for progress supervision.}
\label{tab:app:skill_definitions}
\resizebox{\linewidth}{!}{
\begin{tabular}{c l l l}
\toprule
\textbf{Task} & \textbf{Skill Description} & \textbf{Terminal condition} & \textbf{Progress type} \\
\midrule

\multirow{5}{*}{\rotatebox[origin=c]{90}{\shortstack{Lamp\\replacement}}}
& Reach bulb
& End-effector reaches the bulb region
& Cont. \contprog \\

& Grasp bulb
& Stable grasp is established on the bulb
& Cont. \contprog\\

& Reach lamp socket
& End-effector reaches the socket region while holding the bulb
& Cont. \contprog\\

& Align bulb
& Bulb is aligned with the socket opening
& Disc. \discprog\\

& Twist bulb
& Bulb reaches the end of the thread, producing increased tactile force
& Disc. \discprog\\

\midrule

\multirow{5}{*}{\rotatebox[origin=c]{90}{\shortstack{Bottle\\capping}}}
& Reach bottle cap 
& End-effector reaches the target object 
& Cont. \contprog\\

& Grasp cap 
& Stable grasp is established on the cap 
& Cont. \contprog \\

& Reach bottle 
& End-effector reaches the bottle opening while holding the cap 
& Cont. \contprog\\

& Align cap 
& Cap is aligned with the bottle opening 
& Disc. \discprog\\

& Slide cap 
& Cap slides against the bottle until slippage is detected 
& Disc. \discprog\\

\midrule

\multirow{5}{*}{\rotatebox[origin=c]{90}{\shortstack{Blind\\cube search}}}
& Reach box $k$ 
& End-effector reaches the candidate box region 
& Cont. \contprog \\

& Inspect box $k$ 
& Grasp force reaches a level at which the cube can be identified or ruled out
& Cond. \condprog\\

& Withdraw 
& End-effector leaves the interaction region before reaching box $k{+}1$
& Cont. \contprog\\

& Reach puzzle box 
& End-effector reaches the puzzle box region while holding the cube
& Cont. \contprog \\

& Insert cube 
& Cube is inserted into the square hole 
& Cont. \contprog \\

\bottomrule
\end{tabular}
}
\end{table}

\begin{figure}[h!]
  \centering
  \includegraphics[width=0.9\linewidth]{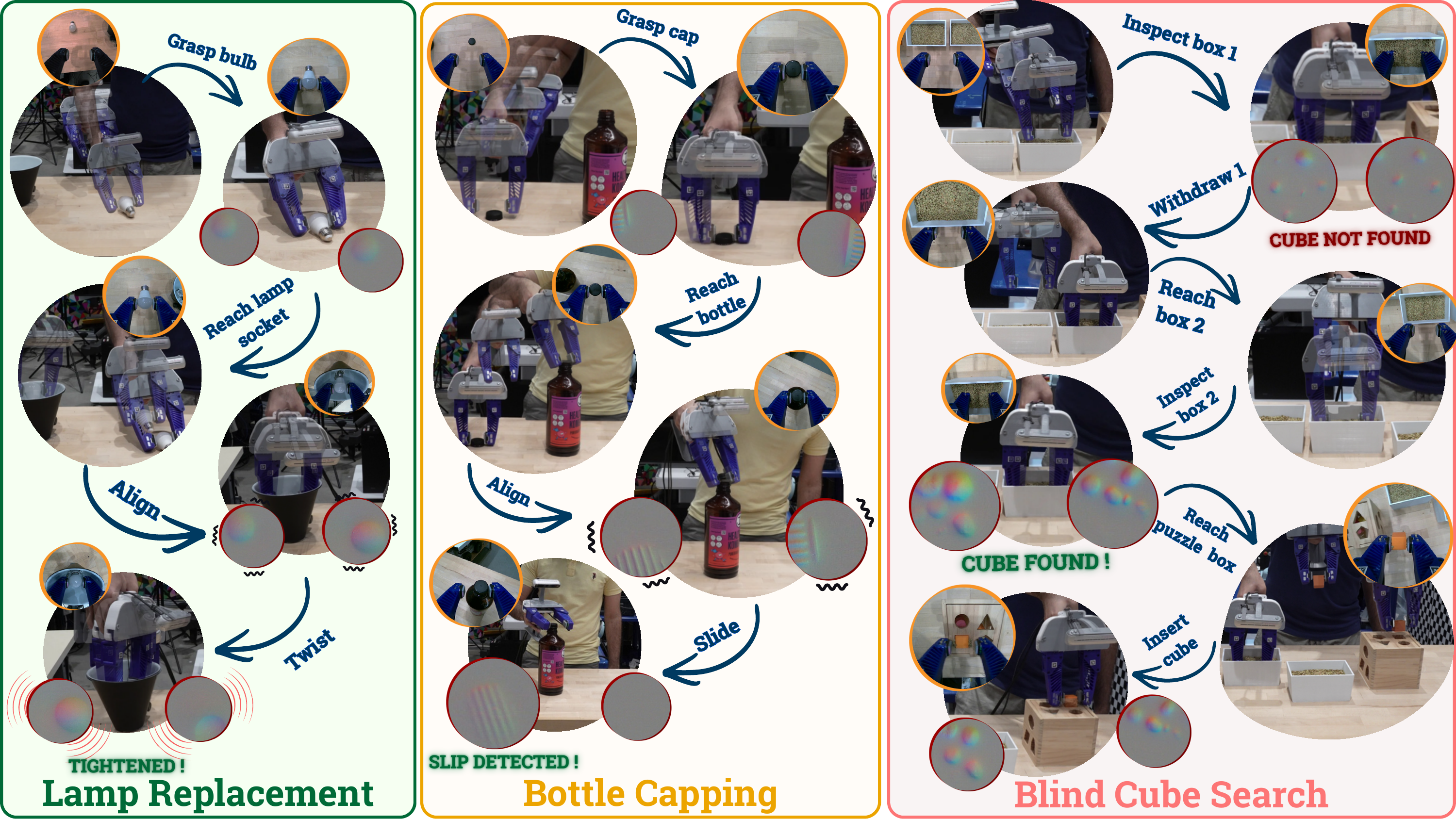}
  \vspace{-0.2cm}
\caption{\textbf{Contact-rich manipulation tasks and their representation requirements.}
\emph{Left:} Lamp replacement involves grasping, socket alignment, and twisting
until tightened. \emph{Middle:} Bottle capping involves grasping, thread
alignment, and sliding, with slip providing a salient tactile event.
\emph{Right:} Blind cube search involves successive inspections and withdrawal
after an unsuccessful attempt, followed by insertion into a puzzle box once
the target is found. Arrows indicate primitive skills. Large images highlight a task execution; orange-bordered insets show egocentric RGB views, and red-bordered insets show tactile readings from the two fingers. Together, the tasks
illustrate skill-specific tactile patterns and complementary needs for 
sensory history and memory of prior interactions for estimating their progress.}
  \label{fig:tasks}
\end{figure}

\paragraph{Single-stream versus cross-attention fusion.}
Our choice of bidirectional self-attention over a unified token sequence is related to the distinction between single-stream and dual-stream vision-language architectures. In dual-stream or cross-attention formulations, language is often used as a query over sensory features. This can be effective for selecting or gating relevant visual tokens, but the resulting representation is still primarily an attended summary of the sensory stream. The language condition may not be explicitly preserved throughout later layers.

This distinction is important for skill progress estimation. In our setting, progress is not a function of the observation alone, but of both the observation and the queried skill. The same RGB and tactile observations can correspond to very different progress values depending on the skill query. For example, a frame near the bottle opening may indicate high progress for \emph{reach bottle}, but low progress for \emph{slide cap}; similarly, a tactile contact pattern may be relevant for \emph{grasp cap} but not for \emph{reach bottle}. Therefore, language should not only gate sensory features, but remain part of the representation used to compute progress.

\AlgName{} addresses this by treating the skill description as a first-class token in the same sequence as RGB, tactile, event-memory, and output tokens.
Self-attention allows language, current observations, event memory, and the output token to mutually update each other across layers. This enables the output representation to remain explicitly skill-conditioned, which is especially important when visually similar observations occur across adjacent skills or when tactile observations must be interpreted differently depending on the queried primitive.

\section{Experiments Details}

We compare \AlgName{} against several baselines that isolate the effect of language conditioning, temporal horizon, tactile sensing, and event memory. We use a batch size of 45, train for 100 epochs, use AdamW with base learning rate \(10^{-4}\), and fine-tune pretrained encoders with a smaller learning rate of \(10^{-5}\). All transformer-based models use hidden dimension \(d=512\), 8 attention heads, and 4 transformer layers. We use active-skill sampling with \(\alpha=0.8\), so most training queries correspond to the currently active skill while a small fraction are sampled from other skills to enforce language conditioning.

\subsection{Baselines Explained}
\paragraph{FiLM-conditioned baseline.}
The FiLM baseline replaces the self-attention-based language-conditioned representation with a FiLM-style conditioning mechanism. It uses the same RGB and tactile horizons as the main model, with \(\Hrgb=2\) and \(\Htac=6\), but does not use event memory. The architecture conditions visual and tactile features on the queried skill embedding through feature-wise affine modulation before progress estimation. This baseline tests whether explicit language modulation through FiLM is sufficient for progress estimation, compared to our token-based formulation that jointly attends over language, RGB, tactile, and memory tokens.

\paragraph{\AlgName{} (h=2) baseline.}
The short-horizon baseline uses the same architecture as \AlgName{} but restricts both RGB and tactile inputs to two frames, i.e., \(\Hrgb=\Htac=2\). Event memory and tactile inputs are still enabled. This baseline reflects the common observation formulation used in manipulation policies, where the model receives only a small fixed window of recent observations. Comparing against this baseline tests whether modality-specific temporal horizons are necessary, especially for tactile signals whose informative patterns often emerge over multiple frames.

\paragraph{Vision only baseline.}
The Vision-only baseline removes both tactile streams and estimates progress using only RGB observations and the queried skill description. This baseline measures how much progress estimation can be solved from visual information alone. It is particularly important for contact-rich skills such as \emph{twist bulb}, \emph{slide cap}, and \emph{inspect box}, where the terminal condition depends on contact force, slippage, or tactile shape information rather than appearance.

\paragraph{No FAF encoder baseline.}
The generic tactile encoder baseline keeps the same architecture and tactile inputs as \AlgName{}, but replaces the tactile-specialized FAF encoder with a generic ImageNet-pretrained DINOv2 encoder. This baseline isolates the importance of tactile pretraining. Since tactile images differ significantly from natural images, this comparison tests whether contact-aware tactile features are needed for reliable progress estimation.

\paragraph{No event memory baseline.}
The no-event-memory baseline removes the event-memory tokens while keeping the current RGB, tactile, and language tokens unchanged. This baseline tests whether storing the terminal observation of the previous skill helps reduce ambiguity near skill transitions. Without event memory, the model must infer the current phase only from the recent observation window, which can be ambiguous when adjacent skills have visually similar states but different completion criteria.

\paragraph{Robometer fine-tuning.}
We start from the released Robometer-4B checkpoint~\cite{robometer2025}, which builds on
Qwen3-VL-4B-Instruct and was pretrained on RBM-1M, and adapt it to our tasks with LoRA (rank~32,
$\alpha{=}64$, dropout~0.05) on all attention and MLP projections of the language model, while training
its progress, success and preference heads in full. Fine-tuning uses the bottle capping and lamp
replacement training demonstrations, converted to Robometer's format with per-frame progress labels for
each subtask, and mixes progress and preference samples as in the original recipe. Each sample contains
48 frames drawn from the episode. We train with a global batch of 8 on two H-200 GPUs for up to 3{,}000 steps,
using AdamW with a learning rate of $2{\times}10^{-5}$, 10\% warmup and a cosine schedule, and keep the
checkpoint with the best held-out progress correlation.

\paragraph{ReWiND training.}
ReWiND~\cite{rewind2025} predicts progress with a four-layer transformer (hidden size~512, 8~heads) over
frozen DINOv2 ViT-B/14 image features and MiniLM-L12 language embeddings. We train a single model jointly
on our bottle capping and lamp replacement demonstrations and on Open X-Embodiment, with 30\% of each
batch drawn from our data. Following the original method, 80\% of training clips are augmented with video
rewinding, which appends a reversed segment so that the model learns progress can decrease, and 20\% pair
a clip with a mismatched instruction as negatives with zero progress. Clips contain up to 16 frames, and
we train for 20{,}000 steps with a batch of 256 and a learning rate of $10^{-4}$ decaying to $10^{-5}$
after 500 warmup steps, selecting the checkpoint with the best held-out per-frame correlation.

\section{Further Discussions on Robustness}
\paragraph{What the perturbations test.}
Figure~\ref{fig:robustness} examines whether the representations
remain useful when image appearance, sensor correspondence, temporal
sampling, or access to observations changes at test time. These
interventions probe different dependencies: appearance tests target
visual cues, geometric tests disrupt spatial relationships, temporal
tests alter motion information, and input removal tests reliance on
touch and memory. All MSE values below are reported in $\times10^{-3}$.

\paragraph{Brightness tests sensitivity to image intensity.}
Darkening by factors of $0.75$ and $0.5$ tests whether predictions
depend on the image intensity seen during training. Unlike a change
in task state, this transformation preserves the scene layout while
reducing the visibility of image details. At half brightness,
\AlgName{}'s MSE changes from $3.4$ to $3.8$ on lamp replacement
and from $4.6$ to $4.7$ on bottle capping, but rises from $15.1$
to $25.3$ on blind search. This suggests greater sensitivity to
visual appearance when interpreting the repeated interactions and
spatial context of the search task.

\paragraph{Hue shifts test dependence on color.}
Hue shifts from $18^\circ$ to $108^\circ$ alter color while
preserving spatial arrangement, testing whether predictions depend
on familiar colors rather than task-relevant geometry and contact.
At $108^\circ$, lamp and bottle MSE remain low at $4.2$ and $5.1$,
respectively, whereas blind-search MSE rises to $54.4$. The assembly
results suggest that exact color is not essential to their progress
predictions. The search result reveals sensitivity to altered visual
appearance, but does not by itself establish that the model uses
color to identify boxes.

\paragraph{Blur tests dependence on fine image detail.}
Gaussian blur with kernels from $3$ to $15$\,px suppresses edges
and texture, probing whether the model requires precise image
details to interpret manipulation. At $15$\,px, bottle-capping MSE
remains close to clean performance ($4.7$ versus $4.6$), whereas
lamp MSE rises from $3.4$ to $6.8$ and blind-search MSE from
$15.1$ to $54.4$. FiLM's bottle-capping MSE already rises from
$11.5$ to $30.9$ at $7$\,px. The differing responses are consistent
with complementary sensing requirements: contact observations can
help resolve cap sliding, while visual detail remains useful for
alignment and interpreting search interactions.

\paragraph{Sampling speed tests tolerance to temporal scale.}
The speed perturbation multiplies the spacing between sampled
frames by $0.5$ or $2$, changing the elapsed span of an observation
window without changing its number of timesteps. It therefore tests
sensitivity to temporal sampling rather than the physical effects
of executing the robot faster or slower. \AlgName{}'s MSE changes
by at most $3.0$ across the three tasks, suggesting that its
predictions tolerate these changes in the sampled rate of motion
and contact evolution.

\paragraph{Frame reversal tests the importance of temporal order.}
Reversal preserves the images in each window but presents them in
the opposite order. This distinguishes sensitivity to the sequence
of observations from dependence on their presence alone. For
example, a developing contact followed by slip conveys different
information from the reversed sequence, even though the individual
images are unchanged. Reversal increases \AlgName{}'s MSE from
$3.4$ to $6.6$ on lamp replacement and from $4.6$ to $11.7$ on
bottle capping, compared with only $15.1$ to $16.2$ on blind search.
The larger assembly effects support the importance of ordered
contact dynamics in these tasks.

\paragraph{Geometric perturbations probe visual--tactile correspondence.}
H-flip mirrors the wrist-camera image and both tactile images while
keeping each tactile stream in its original input slot. This changes
the apparent locations of the fingers and contacts without exchanging
their sensor identities. H-flip + swap additionally exchanges the
tactile streams, accounting for the reversal of finger positions.
Comparing these interventions probes whether the representation
depends on correspondence between visually observed contact and
finger-specific tactile patterns. However, image transformations
alone do not guarantee a physically equivalent mirrored interaction,
and their effects do not directly identify a particular attention
mechanism.

\paragraph{Bottle capping depends on finger-specific contact patterns.}
During cap alignment and repetitive sliding, contact can develop
differently across the two fingers. The location and evolution of
these contacts can help distinguish continued sliding from the slip
event preceding the final twist. Mirroring images without exchanging
sensor identities can disrupt the learned association between these
tactile patterns and their visual locations. Consistent with this
interpretation, \AlgName{}'s MSE increases from $4.6$ to $68.4$
under H-flip, then falls to $12.4$ with the additional swap.
The substantial recovery supports sensitivity to sensor correspondence,
while the remaining gap indicates that swapping does not remove
all effects of mirroring.

\paragraph{Lamp replacement shows a different sensitivity.}
Bulb tightening involves a grasp in which increasing resistance
can affect both tactile sensors. If completion cues are shared
across the fingers, their exact left--right assignment may matter
less than their evolution over time. H-flip increases MSE from
$3.4$ to $8.9$, and swapping the tactile streams leaves it at
$8.9$. This is consistent with disruption of visual geometry or
within-sensor contact patterns that exchanging streams cannot
correct; it does not establish that the two fingers experience
identical forces.


\paragraph{Tactile blackout tests dependence on contact observations.}
Blacking out tactile images removes contact information from both
current and retained observations. This tests whether the model
uses touch when visually similar motions have different physical
outcomes. Lamp MSE rises from $3.4$ to $17.8$, and bottle MSE
from $4.6$ to $63.4$, consistent with the importance of resistance
during tightening and slip during cap sliding. Blind-search MSE
changes little ($15.1$ to $14.2$), indicating limited dependence
on tactile input in this evaluation. Because blackout introduces
an unfamiliar input, it measures sensitivity to removing touch
at deployment rather than performance after retraining without it.

\paragraph{Event blackout tests dependence on retained history.}
Blacking out event memory removes access to earlier interactions
while preserving the current observation window. Lamp and bottle
MSE rise to $6.4$ and $13.2$, respectively, while blind-search
MSE rises from $15.1$ to $171.4$. This task requires remembering
inspection outcomes to interpret box-specific skill progress:
similar current observations may belong to different branches
depending on where the cube was found. The large degradation
supports the contribution of retained history beyond local sensing.
This intervention complements \woevent{}, which is trained to
operate without memory rather than having memory removed at test time.

\paragraph{Event jitter tests sensitivity to retention timing.}
Perturbing detected event timestamps by $\pm5$ frames tests whether
memory depends on selecting an exact observation at a skill
transition. Performance changes little across the three tasks,
suggesting tolerance to small localization errors. A plausible
explanation is that each memory entry contains an observation
window: nearby event timestamps can retain overlapping observations,
and some contact outcomes remain visible beyond a single frame.
This result concerns small timing offsets, rather than missing
events or incorrect event identities.

\paragraph{Memory complements the current observation window.}
Without events, the model must infer progress entirely from its
recent observations. Frame reversal produces larger error increases
for \woevent{} on both assembly tasks, consistent with greater
dependence on local temporal patterns. Retained events provide
additional context after informative interactions leave that window.
This benefit is task-dependent: memory improves robustness in
several settings, but does not prevent failures when geometric
perturbations disrupt the sensory relationships the model has learned.

\section{Limitations.}
Our main focus in this work was the representation quality that we assessed through progress estimation. However, our current formulation assumes access to a skill graph of predefined language-described skills and annotated skill boundaries for progress supervision. While this allows controlled analysis of representation quality, future work should learn skill segmentation and progress labels with weaker supervision. Integrating \AlgName{} into downstream control policies and measuring improvements in task success remains an important next step. Additionally, All experiments use the same sensing platform and evaluate unseen episodes from the three collected task domains. Our results therefore establish robustness across different skill types and sensory dependencies within this setting, but do not establish transfer to unseen object categories, environments, embodiments, or sensor configurations. Evaluating such transfer is an important next step.

\end{document}